\documentclass[sigconf]{acmart}

\setcopyright{none}
\renewcommand\footnotetextcopyrightpermission[1]{}
\setcopyright{none}
\makeatletter
\renewcommand\@formatdoi[1]{\ignorespaces}
\makeatother

\usepackage{balance}
\usepackage{multirow}
\AtBeginDocument{%
  \providecommand\BibTeX{{%
    \normalfont B\kern-0.5em{\scshape i\kern-0.25em b}\kern-0.8em\TeX}}}

\begin{document}

\title{Scalable Cross-lingual Document Similarity through Language-specific Concept Hierarchies}

\author{Carlos Badenes-Olmedo}
\email{cbadenes@fi.upm.es}
\orcid{0000-0002-2753-9917}
\affiliation{%
  \institution{Ontology Engineering Group, Universidad Polit\'ecnica de Madrid}
  \city{Boadilla del Monte}
  \state{Spain}
}

\author{Jos\'e Luis Redondo-Garc\'ia}
\email{jluisred@amazon.com}
\orcid{0000-0002-7413-447X}
\affiliation{%
  \institution{Amazon Research}
  \city{Cambridge}
  \state{UK}
}

\author{Oscar Corcho}
\email{ocorcho@fi.upm.es}
\orcid{0000-0002-9260-0753}
\affiliation{%
  \institution{Ontology Engineering Group, Universidad Polit\'ecnica de Madrid}
  \city{Boadilla del Monte}
  \state{Spain}
}

\renewcommand{\shortauthors}{Badenes-Olmedo C., et al.}

\begin{abstract}
With the ongoing growth in number of digital articles in a wider set of languages and the expanding use of different languages, we need annotation methods that enable browsing multi-lingual corpora. Multilingual probabilistic topic models have recently emerged as a group of semi-supervised machine learning models that can be used to perform thematic explorations on collections of texts in multiple languages. However, these approaches require theme-aligned training data to create a language-independent space. This constraint limits the amount of scenarios that this technique can offer solutions to train and makes it difficult to scale up to situations where a huge collection of multi-lingual documents are required during the training phase. This paper presents an unsupervised document similarity algorithm that does not require parallel or comparable corpora, or any other type of translation resource. The algorithm annotates topics automatically created from documents in a single language with cross-lingual labels and describes documents by hierarchies of multi-lingual concepts from independently-trained models. Experiments performed on the English, Spanish and French editions of JCR-Acquis corpora reveal promising results on classifying and sorting documents by similar content.
\end{abstract}



\maketitle

\section{Introduction}

Cross-language information extraction deals with the retrieval of documents written in languages different from the language of the user's query. At execution time, the query in the source language is typically translated into the target language of the documents with the help of a dictionary or a machine-translation system. But for many languages we may not have access to translation dictionaries or a full translation system, or they can be expensive to apply in an online search system. In such situations it is useful to rely on smaller annotation units derived from the text so the full content doesn't need to be translated, for instance by finding correspondences with regard to the topics discussed. In this case, it may be advisable to automatically learn cross-lingual topics to browse multi-lingual document collections.

Multi-lingual topic models discover language-specific descriptions of each topic from documents in multi-lingual corpora. They are mainly based on Latent Dirichlet Allocation (LDA) \cite{Blei2003}, adding supervised associations between languages by using \textit{parallel} corpus, with sentence-aligned documents (e.g. Europarl\footnote{https://ec.europa.eu/jrc/en/language-technologies/dcep} corpora), or \textit{comparable} corpus, with theme-aligned documents (e.g. Wikipedia\footnote{https://www.wikipedia.org/} articles), in multiple languages. These requirements restrict the kind of corpora that can be used for training since large parallel corpora are rare in most of the use cases, especially for languages with fewer resources. Wikipedia, for example, contains texts in 304 languages but 255 of them have less than 3\% of articles\footnote{https://meta.wikimedia.org/wiki/List\_of\_Wikipedias}. Therefore, the requirement of parallel/comparable corpora for multilingual topic models limits their usage in many situations. In addition, these models rely on associations between documents prior to training. So in order to incorporate new languages or update the existing associations, the model must be re-trained with documents from all languages, making it difficult to scale to large corpora \cite{Hao2018} \cite{Moritz2017}.

\begin{figure*}
\includegraphics[width=13cm,height=8cm]{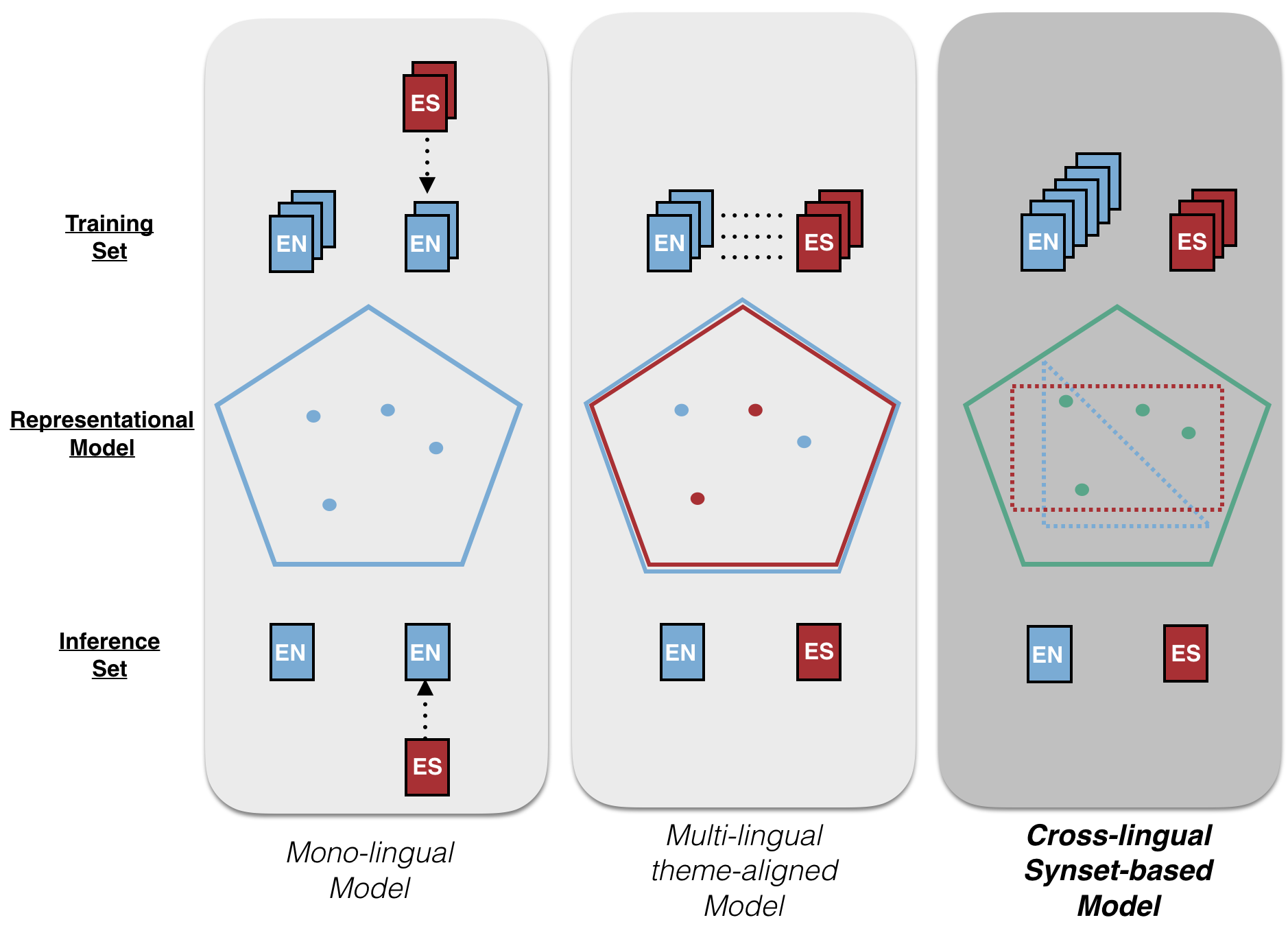}
\caption{Graphical representation of the model that relies on the latent layer of cross-lingual topics obtained by LDA and hash functions through hierarchies of synsets. Mono-lingual approaches force to translate the documents to the same language to represent them in a unique feature space. Multi-language approaches require previously aligned topics from different languages so that documents can be represented in an equivalent feature space. Cross-lingual Synset-based approach creates a new space by combining the feature spaces of each language (i.e synsets from topn topic words). Documents are then represented in this unique space.}
\label{fig:hash_functions}
\end{figure*}

Another approach is to use multi-lingual dictionaries as supervised methods \cite{Ma2017}\cite{errez2016}. They are usually easier to obtain and more widely available than parallel corpora (e.g. PANLEX\footnote{https://panlex.org} covers 5,700 languages and Wiktionary\footnote{https://www.wiktionary.org} covers 8,070 languages). Models built on dictionaries rather than a parallel/comparable corpora are potentially applicable to many more use cases. And even coherent multi-lingual topics can be learnt from partially and fully incomparable corpora with limited amounts of dictionary resources \cite{Hao2018b}.

But all these probabilistic topic models are based on prior knowledge. Connections at document-level (by parallel or comparable corpora) or at word-level (by dictionaries) are created in the training-data before building the model. In this way, the pre-established language relations condition the creation of the topics (supervised method), instead of being inferred from the topics themselves as a posteriori knowledge (non-supervised method). We propose a completely unsupervised way of building cross-lingual topic models that uses sets of cognitive synonyms (synsets) to establish relations between language-specific topics once the model is created and does not require parallel or comparable data for training. These models can be used for large-scale multi-lingual (1) document classification and (2) information retrieval tasks. Our main contributions, described in this paper, are:
\begin{itemize}
    \item a novel \textbf{cross-lingual document similarity algorithm} based on hierarchies of synsets. 
    \item an open-source \textbf{implementation} \footnote{https://github.com/cbadenes/crosslingual-semantic-similarity} of the algorithm
    \item \textbf{data-sets} and \textbf{pre-trained models} to facilitate other researchers to replicate our experiments and validate and test their own ideas.
\end{itemize}

\section{Related Work}

One of the greatest advantages of using probabilistic topic models (PTM) in large document collections is their ability to represent documents as probability distributions over a fixed number of topics, thereby mapping documents into a low-dimensional latent space (the $K$-dimensional probability simplex, where $K$ is the number of topics). A document, represented as a point in this simplex, is considered to have a particular topic distribution. This brings a lot of potential when applied over different information-retrieval (IR) tasks, as evidenced by recent works in different domains such as scholarly \cite{Gatti2015}, health \cite{Lu2016} \cite{TapiNzali2017}, legal \cite{ONeill2017}\cite{Greene2016}, news \cite{He2017} and social networks \cite{Ramage2010} \cite{Cheng2014a}. 

Multilingual probabilistic topic models (MuPTM) \cite{Vulic2015} have recently emerged as a group of language-independent generative machine learning models that can be used on large-volume theme-aligned multilingual text. Due to its generic language-independent nature and the power of inference on unseen documents, MuPTM's have found many interesting applications in many different cross-lingual tasks. They have been used on cross-lingual event clustering \cite{DeSmet2009}, document classification \cite{10.1007/978-3-642-20841-6_45} \cite{Ni:2011:CLT:1935826.1935887},  semantic similarity of words \cite{Mimno:2009:PTM:1699571.1699627}  \cite{Vulic:2012:DHC:2380816.2380872}, information retrieval \cite{10.1007/978-3-642-36973-5_9} \cite{ganguly-etal-2012-cross}, document matching \cite{Platt:2010:TDR:1870658.1870683} \cite{zhu-etal-2013-building}, and others. 

Once a PTM or MuPTM has been generated, documents can be represented by data points in a feature space based on topics to detect similarities among them exploiting inference results and using distance calculation metrics on it. Since exact similarity computations are unaffordable for neighbours detection tasks ($O(n^2)$), some algorithms based on approximate nearest neighbor (ANN) techniques have been proposed to efficiently perform document similarity search based on the low-dimensional latent space created by probabilistic topic models\cite{Zhen2016} \cite{Mao2017}. They transform data points from the original feature space into a hash-code space, so that similar data points have larger probability of collision (i.e. having the same hash code). However, the smaller space created by existing hashing methods lose the exploratory capabilities that topic models offer and the explanatory power that topics have to support the document similarity. The notion of topics is discarded and therefore the ability to make thematic explorations of documents. Recently, a hashing algorithm that groups similar documents and preserves the notion of topics has been proposed \cite{Badenes-Olmedo2019}. It defines a hierarchical set-type data where each level of the hierarchy indicates the importance of the topic according to its distribution. Level 0 contains the topics of the document with the highest score. Level 1 contains the topics with highest score once the first ones have been eliminated, and so on. The knowledge provided by the topics to describe the documents is maintained and an efficient exploration of document collections on a large scale can be performed.

In this paper we take these hierarchies of PTM a step further, to make them cross-lingual. Documents from multi-language corpora can then be efficiently browsed and related without the need for translation. An algorithm that annotates topics with knowledge from a lexical data base and describes documents with hierarchical expressions of multi-lingual concepts is presented in this paper. Hash codes are created from those concept hierarchies to perform document classification and information retrieval tasks on large document collections.

\section{An Approach to Calculate Cross-lingual Document Similarity Efficiently}

Our similarity algorithm considers that cross-lingual models can be built from non-parallel or even non-comparable collections of multi-lingual documents. It first creates a probabilistic topic model for each language separately, and then annotates the topics with cross-lingual labels (Fig \ref{fig:hash_functions}). In the same way, the topic distribution of documents expressed through weighted vectors are first transformed into hierarchies of topics according to their relevance. And then documents are described by a 3-level hierarchy of cross-lingual concepts.


\begin{figure}[t]
\includegraphics[scale=0.3]{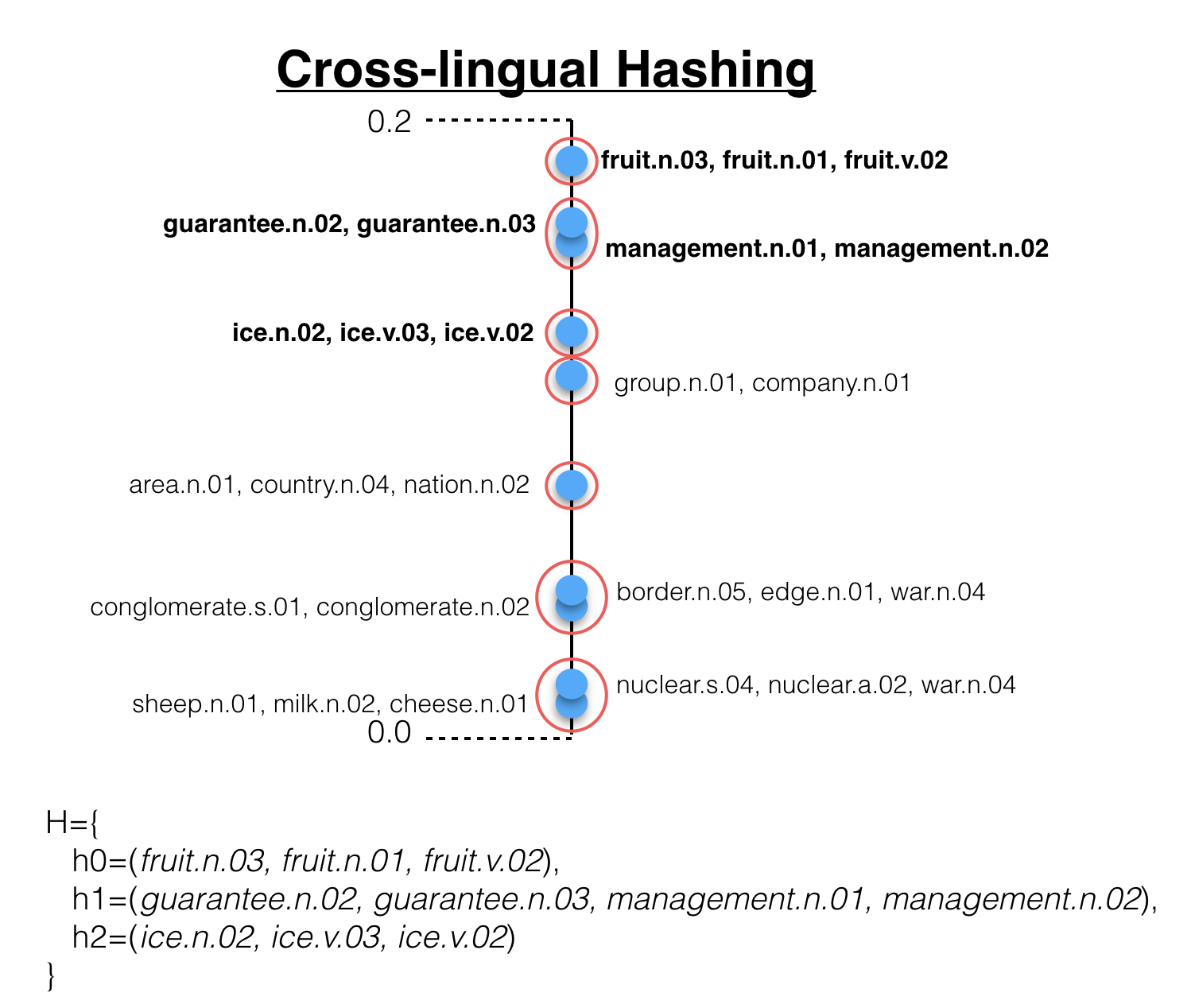}
\caption{Cross-lingual hash-expression (H) of a document based on WordNet-synset annotations created from the top words of each topic distribution. The most relevant topics are grouped according to their importance in three levels (h0, h1 and h2)}
\label{fig:density_hash}
\end{figure}

\subsection{Synset-based annotations}


Each topic is annotated with a list of synset \cite{Bond2013} retrieved from WordNet\footnote{https://wordnet.princeton.edu/}\cite{Miller1995WordNet:English} based on its top$n$ words (Fig \ref{fig:density_hash}). Word by word are queried in WordNet to retrieve its synsets. The final set of synsets for a topic is the union of the synsets from the individual top-words of a topics. Based on empirical evidence from different executions of the algorithm, n=5 is the configuration that offered the best performance in our tests. Let's look at an example to clarify how it works. Given the topics of Table \ref{tb:topics}, the EN-Topic (\textit{"communications systems"}) is annotated with the following synset list: \textit{radio.a.01, radio.v.01, radio.n.03, radio.n.01, radio\_receiver.n.01, equipment.n.01, network.n.02, network.n.04, network.v.01, network.n.05, network.n.01, net.n.06, communication.n.02, communication.n.03, communication.n.01, regulative.s.01}. The list of synset for the ES-Topic (\textit{"sistema de comunicaci\'on"}) is:  \textit{kit.n.02,team.n.01, equipment.n.01, net.n.02, net.n.05, network.n.05, web.n.06, network.n.01, web.n.02, communication.n.02, communication.n.01, announcement.n.02, spectrum.n.02, spectrum.n.01, creep.n.01, ghost.n.01, apparition.n.02, electromagnetic.a.01}. And the list for FR-Topic (\textit{"systeme de communication"}) is:  \textit{access.n.02, approach.n.07, approach.n.02, access.n.06, access.n.03, access.n.05, assault.n.03, bout.n.02, approach.n.01, entree.n.02, entry.n.01, entrance.n.01, entry.n.03, admission.n.01, submission.n.01, introduction.n.01}. The librAIry NLP service\footnote{http://librairy.linkeddata.es/nlp} was used to identify the list of synsets from a topic description based on top words. It includes the Open Multilingual WordNet\footnote{http://compling.hss.ntu.edu.sg/omw/} \cite{Bond2012}.

\subsection{Document representation}
Documents (i.e seen as data points in the generated space) are transformed from the original feature space based on mono-lingual topic distributions into a hierarchical-code space, so that similar data points share relevant cross-lingual concepts. Since topic models create latent themes from word co-occurrence statistics in a corpus, a cross-lingual concept specifies the knowledge about the word-word relations it contains for each language. This abstraction can be extended to cover the knowledge derived from sets of topics. The topics are obtained via state-of-the art methods, collapsed Gibbs sampling\cite{Griffiths2004b} for LDA, and hierarchically divided into groups with different degrees of semantic specificity in a document. Documents represented as a weighted mixture of latent topics (per-document topic distributions) are then annotated in these feature spaces with the relation between topics inside each hierarchy level. Regardless of their language, they are then described by cross-lingual concepts (based on WordNet-synset annotations) and hash codes are calculated to summarize their content \cite{Badenes-Olmedo2019}. The hash expression sets a 3-level hierarchy of cross-lingual concepts. Topics with similar presence in a document are grouped together in the same hierarchical level (Fig \ref{fig:density_hash}). Each level of the hierarchy indicates the importance of the topic according to its distribution. Level 0 describes the topics with the highest score. Level 1 describes the topics with highest score once the first ones have been eliminated, and so on. Documents are described by vectors containing set of topics (i.e. set of synsets), where each dimension means a topic relevance. Given a document $d$ with a topic distribution $q = [t0=0.28, t1=0.05, t2=0.44, t3=0.23]$, the hash expression may be $H_d = {(ts2), (ts0,ts3), (ts1)}$. It means that topic $t2$ described by the synset $ts2$ is the most relevant (i.e 0.44 score), then topics $t0$ and $t3$ described by synsets $ts0$ and $ts3$ (i.e 0.28 and 0.23 scores) and, finally, topic $t1$ described by synset $ts1$ (i.e 0.05).

\begin{figure}[t]
\includegraphics[scale=0.25]{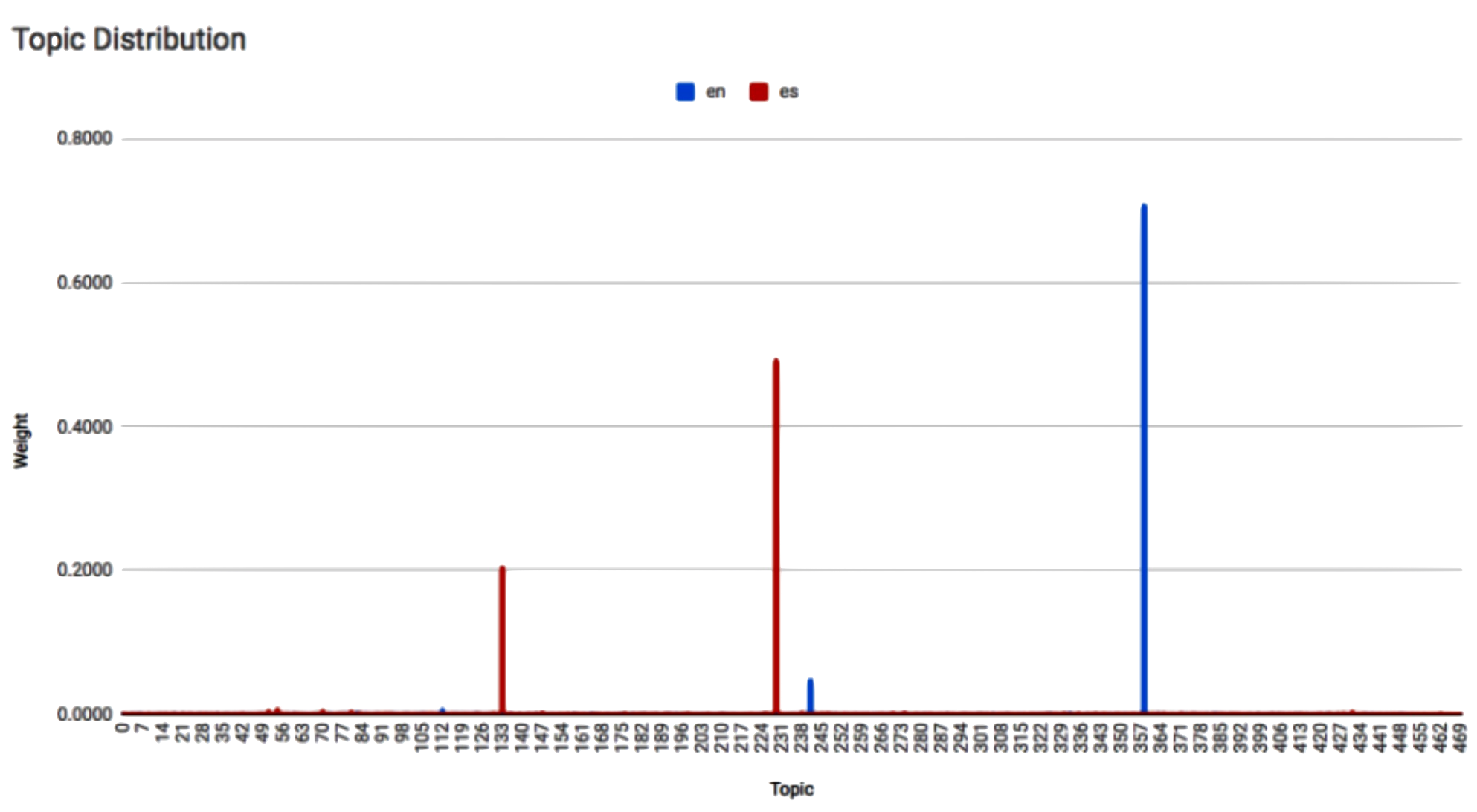}
\caption{topic distributions from the same document in English ($h_{EN}=\{(t3062),(t335),(t8278)\}$) and Spanish ($h_{ES}=\{(t335),(t4060),(t5769)\}$).}
\label{fig:topic_distributions}
\end{figure}

\subsection{Similarity metric}

Since documents are described by set-type data, the proposed distance metric is based on the Jaccard coefficient. This metric is mainly used for this type of data \cite{Li2012} \cite{Ji2013} \cite{Li2010b} \cite{Zhao2013} and computes the similarity of sets by looking at the relative size of their intersection ($\frac{ | A \cap B |}{ | A \cup B |}$). More specifically, the similarity metric used to compare the hash codes created from set of topics is the sum of the Jaccard distances for each hierarchy level, i.e. for each set of topics \cite{Badenes-Olmedo2019}:

\begin{equation}
d_H(H_A,H_B) = \sum\limits_{l=1}^L \Big( d_J(H_A(h_l),H_B(h_l)) \Big) = 
\sum\limits_{l=1}^L \Big( 1 - \frac{H_A(h_l) \cap H_B(h_l)}{H_A(h_l) \cup H_B(h_l)} \Big) 
\label{eq:dh}
\end{equation}

where $H_A$ and $H_B$ are hash codes, $H_A(h_l)$ and $H_B(h_l)$ are the set of topics up to level $l$ for each hash code $H$,  and $L$ is the maximum hierarchy level. A corner case is $L=T$, where $T$ is the number of topics in the model. 

\section{Evaluation}

A way to evaluate our cross-lingual document similarity algorithm is to test how well it performs in practice for different real-life tasks: document classification and information retrieval. Evaluation is done using the B-Cubed metrics \cite{Bagga1998} to estimate the fit between two clusters, the one obtained from a supervised category-based topic alignment algorithm and the one obtained from our unsupervised synset-based topic alignment algorithm. 

Let $CL_i$ be the cluster that document $t_i$ gets clustered in, and $G_i$ its correct cluster from the ground truth. The B-Cubed metric then calculates $precision=\frac{|CL_i \cap G_i|}{|CL_i|}$ and $recall=\frac{|CL_i \cap G_i|}{|G_i|}$. The total precision and recall of the clustering are taken as the average of the precision and recall scores over all documents. Results are also presented in terms of the $F_1$ measure to balance between precision and recall: $F_1=\frac{2 \cdot precision \cdot recall}{precision + recall}$. The aim is to measure the performance of the algorithm taking into account documents with manual category assignments.

\subsection{Data Sets}
A multilingual corpora is required to create the cross-lingual topic models that support our document similarity algorithm. The key feature is that it does not need to be parallel or comparable. However, in order to be able to compare the performance of our unsupervised algorithm with a semi-supervised algorithm (MuPTM-based) it is necessary to use theme-aligned corpora that map topics across languages. We used the JRC-Acquis\footnote{https://ec.europa.eu/jrc/en/language-technologies/jrc-acquis} corpus \cite{Steinberger2006}. It is a collection of legislative texts written in 23 languages, although we only use English, Spanish and French for the tests. Most texts have been manually classified into subject domains according to the EUROVOC\footnote{http://eurovoc.europa.eu/} thesaurus \cite{Eurovoc1995}, which exists in one-to-one translations into approximately twenty languages and distinguishes about 6,000 hierarchically organised descriptors (subject domains). More than 20k documents were used for each language-specific model, a total of 82,140 texts are included in the training-test package, which is publicly available\footnote{http://librairy.linkeddata.es/data/jrc/select?q=*:*} for reuse.


\subsection{Cross-lingual Models}

The JRC-Acquis corpus is annotated with EUROVOC categories. These categories are shared among languages and will serve as support for building the topic models. Moreover, the topic independence assumption \cite{Blei2003} of LDA models should be also satisfied, so the categories must first be moved to their base concepts and therefore disjointed categories. The EUROVOC taxonomy has 7,193 concepts/labels from 21 domain areas such as politics, international relations, european union, law, economics, etc. There are 4,904 reciprocal hierarchical relationships (no polyhierarchy) and 6,992 reciprocal associative relationships. Using hierarchical relations, we identified the root concepts from which all other categories derive. The initial 7,193 labels were then reduced to 452 labels, which are independent (topic independence assumption from LDA is satisfied), and can be used to train the topic models. 

\begin{table*}
  \begin{tabular}{l|l|l}
  \hline
      \multicolumn{1}{c}{EN-Topic 3} & \multicolumn{1}{c}{ES-Topic 3} & \multicolumn{1}{c}{FR-Topic 26}  \\
      \multicolumn{1}{c}{\textit{"communications systems"}} & \multicolumn{1}{c}{\textit{"sistema de comunicaci\'on"}} & \multicolumn{1}{c}{\textit{"systeme de communication"}} \\
  \hline
     radio          & equipo                & communications        \\
     equipment      & red                   & reseaux               \\
     network        & comunicaci\'on        & electroniques          \\
     communication  & espectro              & acces                  \\
     regulatory     & electromagn\'etico    & telecommunications     \\
     spectrum       & electr\'onico         & service                \\
     electronic     & reglamentaci\'on      & universel              \\
     access         & banda                 & reglamentaires         \\
     standard       & etsir                 & nationales             \\
     mobile         & compatibilidad        & fourniture             \\
    \bottomrule
  \end{tabular}
\caption{Randonly selected theme-aligned topics described by top 10 words based on EUROVOC annotations from JRC-Acquis dataset}
\label{tb:topics}
\end{table*}

 A pre-processing of the documents was required to clean texts and to build a suitable data set for the model. We assume that terms with high frequency are not specific to a particular topic, so words present in more than 90\% of the corpus are considered stopwords and removed from the model. Also, rare terms that occur infrequently are considered not representative of a single topic since they do not appear enough to infer that it is salient for a topic. Then, words present in less than 0.5\% of the corpus are also removed from the model. Lemmatized expressions of names, verbs and adjectives were used to create the bag-of-words, and documents with less than 100 characters were discarded since LDA has proven to has lower performance with these type of texts \cite{Cheng2014a}. 
 
 Then, we set the number of topics $K=500$ (several configurations were evaluated, but this was the closest to the performance obtained with the supervised model based on categories). We run the Gibbs samplers for 1000 training iterations on LDA from the open-source librAIry \cite{Badenes-Olmedo2017} software. The Dirichlet priors $\alpha=0.1$ and $\beta=0.01$ were set following \cite{Hu2014a}. Once the word distributions for each topic is available, the list of synsets related with the top5 words for each topic are identified (this number is set to offer better performance after trying several alternatives). Finally, the 3-level hierarchy of topics per document is replaced by a 3-level hierarchy of synsets. Probabilistic topic models in Spanish\footnote{http://librairy.linkeddata.es/jrc-es-model-unsupervised}, English \footnote{http://librairy.linkeddata.es/jrc-en-model-unsupervised} and French\footnote{http://librairy.linkeddata.es/jrc-fr-model-unsupervised} were created independently without previously establishing any type of alignment between their topics.
 
 In order to compare the performance of this non-supervised approach with approaches based on aligned topics, we need to use a variant of LDA to force the correspondence between the 452 root categories identified in the EUROVOC thesaurus and the latent topics of the model. Thus, LabeledLDA \cite{Ramage2009a}, a supervised version of LDA, was used to perform parameter estimation. Theme-aligned probabilistic topic models in Spanish\footnote{http://librairy.linkeddata.es/jrc-es-model}, English \footnote{http://librairy.linkeddata.es/jrc-en-model} and French\footnote{http://librairy.linkeddata.es/jrc-fr-model}) were created sharing the topics but not its definitions (i.e. vocabulary) (see table \ref{tb:topics}).

A simple way of looking at the output quality of the topic models is by simply inspecting top words associated with a particular topic learned during training. A latent topic is semantically coherent if it assigns high probability scores to words that are semantically related \cite{Gliozzo2007} \cite{newman-etal-2010-automatic} \cite{mimno-etal-2011-optimizing}. It is much easier for humans to judge semantic coherence of cross-lingual topics and their alignment across languages when observing the actual words constituting a topic. These words provide a shallow qualitative representation of the latent topic space, and could be seen as direct and comprehensive word-based summaries of a large document collection.

Samples of cross-lingual topics are provided in Table \ref{tb:topics}. We may consider this visual inspection of the top words associated with each topic as an initial qualitative evaluation, suitable for human judges. Documents present similar topic distributions when projecting their content on topics according to their language as can be seen in fig \ref{fig:topic_distributions}. Since the topic identifiers are not aligned, the graphs appear displaced.

\subsection{Cross-lingual Document Classification}
A random group of 1k documents, which have not been used to train the models, is considered for evaluation as they are manually tagged with EUROVOC categories. For each document, the cluster to which it belongs is identified from its categories. This cluster is then compared (B-Cubed metrics) with the one obtained from the labels generated from its most representative topics (\textit{cat}) and with the one obtained from the labels generated with the WordNet-Synsets of those topics (\textit{syn}). Algorithm performance is evaluated in monolingual, bilingual, and multilingual document collections (tables \ref{tb:mono-class} and \ref{tb:multi-class} ) .

\begin{table}[ht]
\begin{center}
\small
\begin{tabular}{cc|rr||rr||rr}
    \hline
    \multicolumn{8}{c}{\textbf{JRC-Acquis Corpora}} \\
    \hline
    & & \multicolumn{2}{c}{\textbf{en}} &
      \multicolumn{2}{c}{\textbf{es}} &
      \multicolumn{2}{c}{\textbf{fr}} \\
    & & {\textit{cat}} & {\textit{syn}} & {\textit{cat}} & {\textit{syn}} & {\textit{cat}} & {\textit{syn}} \\
    \hline
    \multirow{4}{*}{\textbf{prec}} 
    &{\textit{min}}     &0.01 &0.01 &0.01 &0.01 &0.01 &0.01 \\
    &{\textit{max}}     &1.00 &0.95 &1.00 &0.87 &1.00 &0.87 \\
    &{\textit{mean}}    &\textbf{0.58} &0.48 &\textbf{0.55} &0.48 &\textbf{0.55} &0.41 \\
    &{\textit{dev}}     &0.27 &0.23 &0.27 &0.22 &0.26 &0.20 \\
    \hline
    \multirow{4}{*}{\textbf{rec}} 
    &{\textit{min}}     &0.01 &0.03 &0.01 &0.04 &0.01 &0.05 \\
    &{\textit{max}}     &0.96 &1.00 &0.93 &1.00 &0.95 &1.00 \\
    &{\textit{mean}}    &0.39 &\textbf{0.52} &0.36 &\textbf{0.49} &0.42 &\textbf{0.51} \\
    &{\textit{dev}}     &0.24 &0.20 &0.23 &0.20 &0.23 &0.23 \\
    \hline
    \multirow{4}{*}{\textbf{f1}} 
    &{\textit{min}}     &0.02 &0.03 &0.01 &0.02 &0.02 &0.03 \\
    &{\textit{max}}     &0.70 &0.75 &0.70 &0.71 &0.70 &0.73 \\
    &{\textit{mean}}    &0.35 &\textbf{0.42} &0.32 &\textbf{0.41} &0.37 &\textbf{0.39} \\
    &{\textit{dev}}     &0.16 &0.15 &0.15 &0.15 &0.17 &0.17 \\
\end{tabular}
\end{center}
\caption{Document classification performance (precision-'prec', recall-'rec' and fMeasure-'f1') of the categories-based (\textit{cat}) and synset-based (\textit{syn}) topic alignment algorithms in monolingual document collections (en, es, fr)}
\label{tb:mono-class}
\end{table}

\begin{table}[ht]
\begin{center}
\small
\begin{tabular}{cc|rr||rr||rr||rr}
    \hline
    \multicolumn{10}{c}{\textbf{JRC-Acquis Corpora}} \\
    \hline
    & & \multicolumn{2}{c}{\textbf{en-es}} &
      \multicolumn{2}{c}{\textbf{en-fr}} &
      \multicolumn{2}{c}{\textbf{es-fr}} &
      \multicolumn{2}{c}{\textbf{en-es-fr}} \\
    & & {\textit{cat}} & {\textit{syn}} & {\textit{cat}} & {\textit{syn}} & {\textit{cat}} & {\textit{syn}} & {\textit{cat}} & {\textit{syn}} \\
    \hline
    \multirow{4}{*}{\textbf{prec}} 
    &{\textit{min}}     &0.01 &0.01 &0.01 &0.01 &0.01 &0.01 &0.02 &0.02\\
    &{\textit{max}}     &1.00 &0.97 &1.00 &0.98 &1.00 &0.97 &1.00 &0.98\\
    &{\textit{mean}}    &\textbf{0.62} &0.55 &\textbf{0.62} &0.56 &\textbf{0.61} &0.56 &\textbf{0.59} &0.52\\
    &{\textit{dev}}     &0.26 &0.23 &0.25 &0.23 &0.26 &0.23 &0.26 &0.23\\
    \hline
    \multirow{4}{*}{\textbf{rec}} 
    &{\textit{min}}     &0.01 &0.09 &0.00 &0.06 &0.01 &0.07 &0.01 &0.07\\
    &{\textit{max}}     &1.00 &1.00 &0.94 &0.97 &0.91 &0.93 &0.86 &0.93\\
    &{\textit{mean}}    &0.33 &\textbf{0.57} &0.36 &\textbf{0.50} &0.30 &\textbf{0.40} &0.25 &\textbf{0.39}\\
    &{\textit{dev}}     &0.16 &0.23 &0.17 &0.19 &0.13 &0.13 &0.13 &0.15\\
    \hline
    \multirow{4}{*}{\textbf{f1}} 
    &{\textit{min}}     &0.02 &0.02 &0.01 &0.02 &0.02 &0.02 &0.02 &0.05\\
    &{\textit{max}}     &0.75 &0.81 &0.76 &0.81 &0.68 &0.72 &0.62 &0.66\\
    &{\textit{mean}}    &0.36 &\textbf{0.49} &0.38 &\textbf{0.47} &0.35 &\textbf{0.41} &0.30 &\textbf{0.38}\\
    &{\textit{dev}}     &0.16 &0.18 &0.15 &0.18 &0.14 &0.14 &0.11 &0.12\\
\end{tabular}
\end{center}
\caption{Document classification performance (precision-'prec', recall-'rec' and fMeasure-'f1') of the categories-based (\textit{cat}) and synset-based (\textit{syn}) topic alignment algorithms in multi-lingual document collections (en-es, en-fr, es-fr, en-es-fr)}
\label{tb:multi-class}
\end{table}

The results show a higher performance of the semi-supervised algorithm (categories-based topic alignment) in terms of precision, and of the unsupervised algorithm (synset-based topic alignment) in terms of coverage. The cause lies in the set of synonyms generated by WordNet, being able to share the same synset for two different topics. From a more general point of view (fMeasure), the benefit obtained by the increase in coverage (recall) is greater than by the loss of accuracy (precision).

\subsection{Cross-lingual Information Retrieval}
Given a set of documents and a text, the task is to rank the documents according to their relevance to the query text regardless of the language used. The JRC-Acquis corpus is used because by having texts tagged with EUROVOC categories we can build a ground-truth set grouping the documents that share the same codes as those used in the query document. A collection of 1k randomly selected documents (monolingual, bi-lingual and multi-lingual) are annotated by the category-based and synset-based topic alignment algorithms. Then, we randomly take articles to search in D for documents that share the same categories than the query document (i.e  the ground-truth set). Next, the query text is used to search in D for similar documents using category-based annotations and synset-based annotations. We evaluate the performance of the algorithms in terms of precision@3, precision@5 and precision@10 (tables \ref{tb:mono-ir} and \ref{tb:multi-ir} ) .

\begin{table}[ht]
\begin{center}
\small
\begin{tabular}{cc|rr||rr||rr}
    \hline
    \multicolumn{8}{c}{\textbf{JRC-Acquis Corpora}} \\
    \hline
    & & \multicolumn{2}{c}{\textbf{en}} &
      \multicolumn{2}{c}{\textbf{es}} &
      \multicolumn{2}{c}{\textbf{fr}} \\
    & & {\textit{cat}} & {\textit{syn}} & {\textit{cat}} & {\textit{syn}} & {\textit{cat}} & {\textit{syn}} \\
    \hline
    \multirow{2}{*}{\textbf{p@3}} 
    &{\textit{mean}}    &\textbf{0.84} &0.83 &\textbf{0.81} &0.78 &\textbf{0.83} &0.74 \\
    &{\textit{dev}}     &0.26 &0.26 &0.27 &0.29 &0.26 &0.32 \\
    \hline
    \multirow{2}{*}{\textbf{p@5}} 
    &{\textit{mean}}    &\textbf{0.82} &0.80 &\textbf{0.79} &0.75 &\textbf{0.80} &0.72 \\
    &{\textit{dev}}     &0.25 &0.25 &0.25 &0.27 &0.25 &0.29 \\
    \hline
    \multirow{2}{*}{\textbf{p@10}} 
    &{\textit{mean}}    &\textbf{0.77} &0.76 &\textbf{0.75} &0.73 &\textbf{0.77} &0.68 \\
    &{\textit{dev}}     &0.23 &0.25 &0.25 &0.27 &0.24 &0.27 \\
\end{tabular}
\end{center}
\caption{Information retrieval performance (precision@3, precision@5 and precision@10) of the categories-based (\textit{cat}) and synset-based (\textit{syn}) topic alignment algorithms in monolingual document collections (en, es, fr)}
\label{tb:mono-ir}
\end{table}

\begin{table}[ht]
\begin{center}
\small
\begin{tabular}{cc|rr||rr||rr||rr}
    \hline
    \multicolumn{10}{c}{\textbf{JRC-Acquis Corpora}} \\
    \hline
    & & \multicolumn{2}{c}{\textbf{en-es}} &
      \multicolumn{2}{c}{\textbf{en-fr}} &
      \multicolumn{2}{c}{\textbf{es-fr}} &
      \multicolumn{2}{c}{\textbf{en-es-fr}} \\
    & & {\textit{cat}} & {\textit{syn}} & {\textit{cat}} & {\textit{syn}} & {\textit{cat}} & {\textit{syn}} & {\textit{cat}} & {\textit{syn}} \\
    \hline
    \multirow{2}{*}{\textbf{p@3}} 
    &{\textit{mean}}    &\textbf{0.84} &0.79 &\textbf{0.86} &0.77 &\textbf{0.85} &0.78 &\textbf{0.85} &0.75\\
    &{\textit{dev}}     &0.25 &0.28 &0.23 &0.28 &0.25 &0.29 &0.24 &0.31\\
    \hline
    \multirow{2}{*}{\textbf{p@5}} 
    &{\textit{mean}}    &\textbf{0.82} &0.76 &\textbf{0.84} &0.75 &\textbf{0.82} &0.76 &\textbf{0.81} &0.72\\
    &{\textit{dev}}     &0.24 &0.26 &0.23 &0.27 &0.23 &0.27 &0.23 &0.28\\
    \hline
    \multirow{2}{*}{\textbf{p@10}} 
    &{\textit{mean}}    &\textbf{0.78} &0.73 &\textbf{0.80} &0.70 &\textbf{0.77} &0.72 &\textbf{0.76} &0.67\\
    &{\textit{dev}}     &0.22 &0.24 &0.22 &0.24 &0.23 &0.26 &0.23 &0.26\\
\end{tabular}
\end{center}
\caption{Information retrieval performance (precision@3, precision@5 and precision@10) of the categories-based (\textit{cat}) and synset-based (\textit{syn}) topic alignment algorithms in multi-lingual document collections (en-es, en-fr, es-fr, en-es-fr)}
\label{tb:multi-ir}
\end{table}

Although the precision values are lower than those obtained by semi-supervised approximation, they are sufficiently promising (around 0.75) to think that introducing improvements in the lemmatization process would increase the quality of the WordNet-synset annotations derived from the most representative words of each topic (precision values close to 0.8 in the English corpus).

\section{Conclusions}

In this paper we present a first approach towards the calculation of cross-lingual document similarity through unsupervised probabilistic topic models and WordNet-synsets without the need for parallel or comparable corpora.

As expected, the performance of our algorithm in terms of accuracy is not as good as that of the algorithm based on topics previously aligned by documents annotated with categories (theme-aligned training data). However, in terms of coverage, the performance of the unsupervised approach is much greater than that offered by the semi-supervised approach, to the point of offering better overall performance (i.e f1) in classification tasks. 
In addition, the algorithm has proved to perform close to the semi-supervised algorithm in information retrieval task, which makes us think that the process of topic annotation by set of synonyms should be improved to filter those elements that are not sufficiently representative. Our future lines of work will go in that direction, incorporating context information to identify the most representative synset for each topic.

\begin{acks}
This research was partially supported by the European Union's Horizon 2020 research and innovation programme under grant agreement No 780247: TheyBuyForYou, and by the Spanish Ministerio de Econom\'ia, Industria y Competitividad and EU FEDER funds under the DATOS 4.0: RETOS Y SOLUCIONES - UPM Spanish national project (TIN2016-78011-C4-4-R). 
\end{acks}

\bibliographystyle{ACM-Reference-Format}
\bibliography{references}


\begin{thebibliography}{00}


\ifx \showCODEN    \undefined \def \showCODEN     #1{\unskip}     \fi
\ifx \showDOI      \undefined \def \showDOI       #1{#1}\fi
\ifx \showISBNx    \undefined \def \showISBNx     #1{\unskip}     \fi
\ifx \showISBNxiii \undefined \def \showISBNxiii  #1{\unskip}     \fi
\ifx \showISSN     \undefined \def \showISSN      #1{\unskip}     \fi
\ifx \showLCCN     \undefined \def \showLCCN      #1{\unskip}     \fi
\ifx \shownote     \undefined \def \shownote      #1{#1}          \fi
\ifx \showarticletitle \undefined \def \showarticletitle #1{#1}   \fi
\ifx \showURL      \undefined \def \showURL       {\relax}        \fi
\providecommand\bibfield[2]{#2}
\providecommand\bibinfo[2]{#2}
\providecommand\natexlab[1]{#1}
\providecommand\showeprint[2][]{arXiv:#2}

\bibitem[\protect\citeauthoryear{Badenes-Olmedo, Redondo-Garcia, and
  Corcho}{Badenes-Olmedo et~al\mbox{.}}{2017a}]%
        {Badenes-Olmedo2017}
\bibfield{author}{\bibinfo{person}{Carlos Badenes-Olmedo},
  \bibinfo{person}{Jose~Luis Redondo-Garcia}, {and} \bibinfo{person}{Oscar
  Corcho}.} \bibinfo{year}{2017}\natexlab{a}.
\newblock \showarticletitle{{Distributing Text Mining tasks with librAIry}}. In
  \bibinfo{booktitle}{{\em Proceedings of the 17th ACM Symposium on Document
  Engineering (DocEng)}}.
\newblock
\showISBNx{978-1-4503-4689-4/17/09}
\showDOI{%
\url{https://doi.org/10.1145/3103010.3121040}}


\bibitem[\protect\citeauthoryear{Badenes-Olmedo, Redondo-Garcia, and
  Corcho}{Badenes-Olmedo et~al\mbox{.}}{2017b}]%
        {Badenes-Olmedo2017a}
\bibfield{author}{\bibinfo{person}{Carlos Badenes-Olmedo},
  \bibinfo{person}{Jose~Luis Redondo-Garcia}, {and} \bibinfo{person}{Oscar
  Corcho}.} \bibinfo{year}{2017}\natexlab{b}.
\newblock \bibinfo{title}{{librAIry/eval-similarity-calculation}}.
\newblock   (\bibinfo{year}{2017}).
\newblock
\showDOI{%
\url{https://doi.org/10.5281/zenodo.931305}}


\bibitem[\protect\citeauthoryear{Bahmani, Moseley, Vattani, Kumar, and
  Vassilvitskii}{Bahmani et~al\mbox{.}}{2012}]%
        {Bahmani2012}
\bibfield{author}{\bibinfo{person}{Bahman Bahmani}, \bibinfo{person}{Benjamin
  Moseley}, \bibinfo{person}{Andrea Vattani}, \bibinfo{person}{Ravi Kumar},
  {and} \bibinfo{person}{Sergei Vassilvitskii}.}
  \bibinfo{year}{2012}\natexlab{}.
\newblock \showarticletitle{{Scalable K-Means ++}}.
\newblock \bibinfo{journal}{{\em Proceedings of the VLDB Endowment (PVLDB)\/}}
  \bibinfo{volume}{5} (\bibinfo{year}{2012}), \bibinfo{pages}{622--633}.
\newblock
\showISBNx{2150-8097}
\showISSN{2150-8097}
\showDOI{%
\url{https://doi.org/10.14778/2180912.2180915}}


\bibitem[\protect\citeauthoryear{Bart, Welling, and Perona}{Bart
  et~al\mbox{.}}{2011}]%
        {Bart2011}
\bibfield{author}{\bibinfo{person}{Evgeniy Bart}, \bibinfo{person}{Max
  Welling}, {and} \bibinfo{person}{Pietro Perona}.}
  \bibinfo{year}{2011}\natexlab{}.
\newblock \showarticletitle{{Unsupervised organization of image collections:
  Taxonomies and beyond}}.
\newblock \bibinfo{journal}{{\em IEEE Transactions on Pattern Analysis and
  Machine Intelligence\/}} \bibinfo{volume}{33}, \bibinfo{number}{11}
  (\bibinfo{year}{2011}), \bibinfo{pages}{2302--2315}.
\newblock
\showISBNx{1939-3539 (Electronic){\textbackslash}n0098-5589 (Linking)}
\showISSN{01628828}
\showDOI{%
\url{https://doi.org/10.1109/TPAMI.2011.79}}


\bibitem[\protect\citeauthoryear{Blei}{Blei}{2012}]%
        {Blei2012}
\bibfield{author}{\bibinfo{person}{David Blei}.}
  \bibinfo{year}{2012}\natexlab{}.
\newblock \showarticletitle{{Introduction to Probabilistic Topic Modeling}}.
\newblock \bibinfo{journal}{{\it Commun. ACM}}  \bibinfo{volume}{55}
  (\bibinfo{year}{2012}), \bibinfo{pages}{77--84}.
\newblock
\showISBNx{0001-0782}
\showISSN{00010782}
\showDOI{%
\url{https://doi.org/10.1145/2133806.2133826}}


\bibitem[\protect\citeauthoryear{Blei, Carin, and Dunson}{Blei
  et~al\mbox{.}}{2010a}]%
        {Blei2010}
\bibfield{author}{\bibinfo{person}{David Blei}, \bibinfo{person}{Lawrence
  Carin}, {and} \bibinfo{person}{David Dunson}.}
  \bibinfo{year}{2010}\natexlab{a}.
\newblock \showarticletitle{{Probabilistic topic models}}.
\newblock \bibinfo{journal}{{\em IEEE Signal Processing Magazine\/}}
  \bibinfo{volume}{27}, \bibinfo{number}{6} (\bibinfo{year}{2010}),
  \bibinfo{pages}{55--65}.
\newblock
\showISBNx{0805854185}
\showISSN{10535888}
\showDOI{%
\url{https://doi.org/10.1109/MSP.2010.938079}}


\bibitem[\protect\citeauthoryear{Blei, Carin, and Dunson}{Blei
  et~al\mbox{.}}{2010b}]%
        {Blei2010a}
\bibfield{author}{\bibinfo{person}{David Blei}, \bibinfo{person}{Lawrence
  Carin}, {and} \bibinfo{person}{David Dunson}.}
  \bibinfo{year}{2010}\natexlab{b}.
\newblock \showarticletitle{{Probabilistic Topic Models: A focus on graphical
  model design and applications to document and image analysis.}}
\newblock \bibinfo{journal}{{\em IEEE signal processing magazine\/}}
  \bibinfo{volume}{27}, \bibinfo{number}{6} (\bibinfo{year}{2010}),
  \bibinfo{pages}{55--65}.
\newblock
\showISBNx{0805854185}
\showISSN{1053-5888}
\showDOI{%
\url{https://doi.org/10.1109/MSP.2010.938079}}


\bibitem[\protect\citeauthoryear{Blei and Jordan}{Blei and Jordan}{2003}]%
        {Blei2003a}
\bibfield{author}{\bibinfo{person}{David~M Blei} {and}
  \bibinfo{person}{Michael~I Jordan}.} \bibinfo{year}{2003}\natexlab{}.
\newblock \showarticletitle{{Modeling Annotated Data University of California
  Computer Science Division and Department of Statistics University of
  California Modeling Annotated Data}}.
\newblock \bibinfo{journal}{{\em Science\/}} (\bibinfo{year}{2003}).
\newblock


\bibitem[\protect\citeauthoryear{Blei and Lafferty}{Blei and Lafferty}{2007}]%
        {Blei2007a}
\bibfield{author}{\bibinfo{person}{David~M. Blei} {and}
  \bibinfo{person}{John~D. Lafferty}.} \bibinfo{year}{2007}\natexlab{}.
\newblock \showarticletitle{{A correlated topic model of Science}}.
\newblock \bibinfo{journal}{{\em The Annals of Applied Statistics\/}}
  \bibinfo{volume}{1}, \bibinfo{number}{1} (\bibinfo{year}{2007}),
  \bibinfo{pages}{17--35}.
\newblock
\showISBNx{1595933832}
\showISSN{1932-6157}
\showDOI{%
\url{https://doi.org/10.1214/07-AOAS136}}


\bibitem[\protect\citeauthoryear{Blei, Ng, and Jordan}{Blei
  et~al\mbox{.}}{2003}]%
        {Blei2003}
\bibfield{author}{\bibinfo{person}{David~M Blei}, \bibinfo{person}{Andrew~Y
  Ng}, {and} \bibinfo{person}{Michael~I Jordan}.}
  \bibinfo{year}{2003}\natexlab{}.
\newblock \showarticletitle{{Latent Dirichlet Allocation}}.
\newblock \bibinfo{journal}{{\em Journal of Machine Learning Research\/}}
  \bibinfo{volume}{3}, \bibinfo{number}{4-5} (\bibinfo{year}{2003}),
  \bibinfo{pages}{993--1022}.
\newblock
\showISBNx{9781577352815}
\showISSN{15324435}
\showDOI{%
\url{https://doi.org/10.1162/jmlr.2003.3.4-5.993}}


\bibitem[\protect\citeauthoryear{Boyd-Graber and Resnik}{Boyd-Graber and
  Resnik}{2010}]%
        {Boyd-Graber2010}
\bibfield{author}{\bibinfo{person}{Jordan Boyd-Graber} {and}
  \bibinfo{person}{Philip Resnik}.} \bibinfo{year}{2010}\natexlab{}.
\newblock \showarticletitle{{Holistic Sentiment Analysis Across Languages:
  Multilingual Supervised Latent Dirichlet Allocation}}.
\newblock \bibinfo{journal}{{\em Proceedings of the 2010 Conference on
  Empirical Methods in Natural Language Processing\/}}
  \bibinfo{number}{October} (\bibinfo{year}{2010}), \bibinfo{pages}{45--55}.
\newblock
\showISBNx{1932432868}
\showISSN{1098-6596}
\showDOI{%
\url{https://doi.org/10.1017/CBO9781107415324.004}}


\bibitem[\protect\citeauthoryear{Celikyilmaz, Hakkani-Tur, and Tur}{Celikyilmaz
  et~al\mbox{.}}{2010}]%
        {Celikyilmaz2010}
\bibfield{author}{\bibinfo{person}{a Celikyilmaz}, \bibinfo{person}{D
  Hakkani-Tur}, {and} \bibinfo{person}{Gokhan Tur}.}
  \bibinfo{year}{2010}\natexlab{}.
\newblock \showarticletitle{{LDA Based Similarity Modeling for Question
  Answering}}. In \bibinfo{booktitle}{{\em Proceedings of the NAACL HLT 2010
  Workshop on Semantic Search}}. \bibinfo{pages}{1--9}.
\newblock


\bibitem[\protect\citeauthoryear{Dagan, Lee, and Pereira}{Dagan
  et~al\mbox{.}}{1999}]%
        {Dagan1998}
\bibfield{author}{\bibinfo{person}{Ido Dagan}, \bibinfo{person}{Lillian Lee},
  {and} \bibinfo{person}{Fernando C.~N. Pereira}.}
  \bibinfo{year}{1999}\natexlab{}.
\newblock \showarticletitle{{Similarity-Based Models of Word Cooccurrence
  Probabilities}}.
\newblock \bibinfo{journal}{{\em Machine Learning\/}} \bibinfo{volume}{34},
  \bibinfo{number}{1-3} (\bibinfo{year}{1999}), \bibinfo{pages}{43--69}.
\newblock
\showISSN{0885-6125}
\showDOI{%
\url{https://doi.org/10.1023/A:1007537716579}}


\bibitem[\protect\citeauthoryear{Deerwester, Dumais, Furnas, Landauer, and
  Harshman}{Deerwester et~al\mbox{.}}{1990}]%
        {Deerwester1990}
\bibfield{author}{\bibinfo{person}{S Deerwester}, \bibinfo{person}{S~T Dumais},
  \bibinfo{person}{G~W Furnas}, \bibinfo{person}{T~K Landauer}, {and}
  \bibinfo{person}{R Harshman}.} \bibinfo{year}{1990}\natexlab{}.
\newblock \showarticletitle{{Indexing by Latent Semantic Analysis}}.
\newblock \bibinfo{journal}{{\em Journal of the American Society for
  Information Science\/}}  \bibinfo{volume}{41} (\bibinfo{year}{1990}),
  \bibinfo{pages}{391–407}.
\newblock
\showISBNx{9781450300322}
\showISSN{00028231}
\showDOI{%
\url{https://doi.org/10.1002/(SICI)1097-4571(199009)41:6<391::AID-ASI1>3.0.CO;2-9}}


\bibitem[\protect\citeauthoryear{Ester, Kriegel, S, and Xu}{Ester
  et~al\mbox{.}}{1996}]%
        {Ester1996}
\bibfield{author}{\bibinfo{person}{Martin Ester}, \bibinfo{person}{Hans-peter
  Kriegel}, \bibinfo{person}{Jörg S}, {and} \bibinfo{person}{Xiaowei Xu}.}
  \bibinfo{year}{1996}\natexlab{}.
\newblock \showarticletitle{{A density-based algorithm for discovering clusters
  in large spatial databases with noise}}.
\newblock  (\bibinfo{year}{1996}), \bibinfo{pages}{226--231}.
\newblock
\showDOI{%
\url{https://doi.org/citeulike-article-id:3509601}}


\bibitem[\protect\citeauthoryear{Halkidi, Batistakis, and Vazirgiannis}{Halkidi
  et~al\mbox{.}}{2001}]%
        {Halkidi2001a}
\bibfield{author}{\bibinfo{person}{Maria Halkidi}, \bibinfo{person}{Yannis
  Batistakis}, {and} \bibinfo{person}{Michalis Vazirgiannis}.}
  \bibinfo{year}{2001}\natexlab{}.
\newblock \showarticletitle{{On clustering validation techniques}}.
\newblock \bibinfo{journal}{{\em Journal of Intelligent Information Systems\/}}
  \bibinfo{volume}{17}, \bibinfo{number}{2-3} (\bibinfo{year}{2001}),
  \bibinfo{pages}{107--145}.
\newblock
\showISBNx{0925-9902}
\showISSN{09259902}
\showDOI{%
\url{https://doi.org/10.1023/A:1012801612483}}
\showeprint[arxiv]{astro-ph/0005074v1}


\bibitem[\protect\citeauthoryear{Hall, Jurafsky, and Manning}{Hall
  et~al\mbox{.}}{2008}]%
        {Hall2008}
\bibfield{author}{\bibinfo{person}{David Hall}, \bibinfo{person}{Dan Jurafsky},
  {and} \bibinfo{person}{Christopher~D Manning}.}
  \bibinfo{year}{2008}\natexlab{}.
\newblock \showarticletitle{{Studying the History of Ideas Using Topic
  Models}}.
\newblock \bibinfo{journal}{{\em In Proceedings of the 2008 Conference on
  Empirical Methods in Natural Language Processing, Honolulu, October 2008\/}}
  \bibinfo{number}{October} (\bibinfo{year}{2008}), \bibinfo{pages}{363--371}.
\newblock
\showDOI{%
\url{https://doi.org/10.3115/1613715.1613763}}


\bibitem[\protect\citeauthoryear{Hearst and Hall}{Hearst and Hall}{1999}]%
        {Hearst1999}
\bibfield{author}{\bibinfo{person}{Marti~a Hearst} {and} \bibinfo{person}{South
  Hall}.} \bibinfo{year}{1999}\natexlab{}.
\newblock \showarticletitle{{Untangling Text Data Mining}}. In
  \bibinfo{booktitle}{{\em the 37th Annual Meeting of the Association for
  Computational Linguistics}}. \bibinfo{pages}{1--13}.
\newblock


\bibitem[\protect\citeauthoryear{Hofmann}{Hofmann}{2001}]%
        {Hofmann2001}
\bibfield{author}{\bibinfo{person}{Thomas Hofmann}.}
  \bibinfo{year}{2001}\natexlab{}.
\newblock \showarticletitle{{Unsupervised Learning by Probabilistic Latent
  Semantic Analysis}}.
\newblock \bibinfo{journal}{{\em Machine Learning\/}} \bibinfo{volume}{42},
  \bibinfo{number}{1-2} (\bibinfo{year}{2001}), \bibinfo{pages}{177--196}.
\newblock


\bibitem[\protect\citeauthoryear{Jiang, Qian, Shen, Fu, and Mei}{Jiang
  et~al\mbox{.}}{2015}]%
        {Jiang2015}
\bibfield{author}{\bibinfo{person}{Shuhui Jiang}, \bibinfo{person}{Xueming
  Qian}, \bibinfo{person}{Jialie Shen}, \bibinfo{person}{Yun Fu}, {and}
  \bibinfo{person}{Tao Mei}.} \bibinfo{year}{2015}\natexlab{}.
\newblock \showarticletitle{{Author topic model-based collaborative filtering
  for personalized POI recommendations}}.
\newblock \bibinfo{journal}{{\em IEEE Transactions on Multimedia\/}}
  \bibinfo{volume}{17}, \bibinfo{number}{6} (\bibinfo{year}{2015}),
  \bibinfo{pages}{907--918}.
\newblock
\showISBNx{1520-9210 VO - 17}
\showISSN{15209210}
\showDOI{%
\url{https://doi.org/10.1109/TMM.2015.2417506}}


\bibitem[\protect\citeauthoryear{Kenter and Rijke}{Kenter and Rijke}{2015}]%
        {Kenter2015}
\bibfield{author}{\bibinfo{person}{Tom Kenter} {and}
  \bibinfo{person}{Maarten~de Rijke}.} \bibinfo{year}{2015}\natexlab{}.
\newblock \showarticletitle{{Short Text Similarity with Word Embeddings
  Categories and Subject Descriptors}}.
\newblock \bibinfo{journal}{{\em Proceedings of the 24th ACM International on
  Conference on Information and Knowledge Management (CIKM 2015)\/}}
  (\bibinfo{year}{2015}), \bibinfo{pages}{1411--1420}.
\newblock
\showISBNx{9781450337946}
\showDOI{%
\url{https://doi.org/10.1145/2806416.2806475}}


\bibitem[\protect\citeauthoryear{Li, Wang, Lim, Blei, and Fei-fei}{Li
  et~al\mbox{.}}{2010}]%
        {Li2010a}
\bibfield{author}{\bibinfo{person}{Li-jia Li}, \bibinfo{person}{Chong Wang},
  \bibinfo{person}{Yongwhan Lim}, \bibinfo{person}{David~M Blei}, {and}
  \bibinfo{person}{Li Fei-fei}.} \bibinfo{year}{2010}\natexlab{}.
\newblock \showarticletitle{{Building and Using a Semantivisual Image
  Hierarchy}}. In \bibinfo{booktitle}{{\em IEEE Computer Society Conference on
  Computer Vision and Pattern Recognition}}. \bibinfo{pages}{3336--3343}.
\newblock
\showISBNx{9781424469833}


\bibitem[\protect\citeauthoryear{Lin}{Lin}{1991}]%
        {Lin1991}
\bibfield{author}{\bibinfo{person}{Jianhua Lin}.}
  \bibinfo{year}{1991}\natexlab{}.
\newblock \showarticletitle{{Divergence Measures Based on the Shannon
  Entropy}}.
\newblock \bibinfo{journal}{{\em IEEE Transactions on Information Theory\/}}
  \bibinfo{volume}{37}, \bibinfo{number}{1} (\bibinfo{year}{1991}),
  \bibinfo{pages}{145--151}.
\newblock
\showISBNx{0018-9448}
\showISSN{15579654}
\showDOI{%
\url{https://doi.org/10.1109/18.61115}}


\bibitem[\protect\citeauthoryear{Luo, Stenger, Zhao, and Kim}{Luo
  et~al\mbox{.}}{2015}]%
        {Luo2015}
\bibfield{author}{\bibinfo{person}{Wenhan Luo}, \bibinfo{person}{Björn
  Stenger}, \bibinfo{person}{Xiaowei Zhao}, {and} \bibinfo{person}{Tae-Kyun
  Kim}.} \bibinfo{year}{2015}\natexlab{}.
\newblock \showarticletitle{{Automatic Topic Discovery for Multi-Object
  Tracking}}. In \bibinfo{booktitle}{{\em Proceedings of the Twenty-Ninth AAAI
  COnference on Artificial Intelligence}}. \bibinfo{pages}{3820--3826}.
\newblock
\showISBNx{9781577357032}


\bibitem[\protect\citeauthoryear{Pritchard, Stephens, and Donnelly}{Pritchard
  et~al\mbox{.}}{2000}]%
        {Pritchard2000}
\bibfield{author}{\bibinfo{person}{Jonathan~K. Pritchard},
  \bibinfo{person}{Matthew Stephens}, {and} \bibinfo{person}{Peter Donnelly}.}
  \bibinfo{year}{2000}\natexlab{}.
\newblock \showarticletitle{{Inference of population structure using multilocus
  genotype data}}.
\newblock \bibinfo{journal}{{\em Genetics\/}} \bibinfo{volume}{155},
  \bibinfo{number}{2} (\bibinfo{year}{2000}), \bibinfo{pages}{945--959}.
\newblock


\bibitem[\protect\citeauthoryear{Rao}{Rao}{1982}]%
        {Rao1982}
\bibfield{author}{\bibinfo{person}{C~Radhakrishna Rao}.}
  \bibinfo{year}{1982}\natexlab{}.
\newblock \showarticletitle{{Diversity: Its Measurement, Decomposition,
  Apportionment and Analysis}}.
\newblock \bibinfo{journal}{{\em Sankhy{\={a}}: The Indian Journal of
  Statistics, Series A\/}} \bibinfo{volume}{44}, \bibinfo{number}{1}
  (\bibinfo{year}{1982}), \bibinfo{pages}{1--22}.
\newblock
\showISBNx{0581-572X}
\showISSN{0581572X}


\bibitem[\protect\citeauthoryear{Rus, Niraula, and Banjade}{Rus
  et~al\mbox{.}}{2013}]%
        {Rus2013}
\bibfield{author}{\bibinfo{person}{Vasile Rus}, \bibinfo{person}{Nobal
  Niraula}, {and} \bibinfo{person}{Rajendra Banjade}.}
  \bibinfo{year}{2013}\natexlab{}.
\newblock \showarticletitle{{Similarity Measures Based on Latent Dirichlet
  Allocation}}.
\newblock In \bibinfo{booktitle}{{\em Computational Linguistics and Intelligent
  Text Processing}}. \bibinfo{pages}{459--470}.
\newblock


\bibitem[\protect\citeauthoryear{Steyvers, Ths, and {T}}{Steyvers
  et~al\mbox{.}}{2006}]%
        {Steyvers2006}
\bibfield{author}{\bibinfo{person}{M Steyvers}, \bibinfo{person}{Grif Ths},
  {and} \bibinfo{person}{{T}}.} \bibinfo{year}{2006}\natexlab{}.
\newblock \showarticletitle{{Probabilistic topic models}}.
\newblock \bibinfo{journal}{{\em In Landauer, T., McNamara, D., Dennis, S., and
  Kintsch, W., editors, Latent Semantic Analysis: A Road to Meaning. Laurence
  Erlbaum. Tang, Z. and MacLennan, J\/}} (\bibinfo{year}{2006}).
\newblock


\bibitem[\protect\citeauthoryear{Towne, Ros{\'{e}}, and Herbsleb}{Towne
  et~al\mbox{.}}{2016}]%
        {Towne2016}
\bibfield{author}{\bibinfo{person}{W~Ben Towne}, \bibinfo{person}{Carolyn~P
  Ros{\'{e}}}, {and} \bibinfo{person}{James Herbsleb}.}
  \bibinfo{year}{2016}\natexlab{}.
\newblock \showarticletitle{{Measuring Similarity Similarly: LDA and Human
  Perception}}.
\newblock \bibinfo{journal}{{\em ACM Transactions on Intelligent Systems and
  Technology ACM Reference Format ACM Trans. Intell. Syst. Technol\/}}
  \bibinfo{volume}{7}, \bibinfo{number}{2} (\bibinfo{year}{2016}),
  \bibinfo{pages}{1--25}.
\newblock


\end{thebibliography}



\begin{thebibliography}{44}


\ifx \showCODEN    \undefined \def \showCODEN     #1{\unskip}     \fi
\ifx \showDOI      \undefined \def \showDOI       #1{#1}\fi
\ifx \showISBNx    \undefined \def \showISBNx     #1{\unskip}     \fi
\ifx \showISBNxiii \undefined \def \showISBNxiii  #1{\unskip}     \fi
\ifx \showISSN     \undefined \def \showISSN      #1{\unskip}     \fi
\ifx \showLCCN     \undefined \def \showLCCN      #1{\unskip}     \fi
\ifx \shownote     \undefined \def \shownote      #1{#1}          \fi
\ifx \showarticletitle \undefined \def \showarticletitle #1{#1}   \fi
\ifx \showURL      \undefined \def \showURL       {\relax}        \fi
\providecommand\bibfield[2]{#2}
\providecommand\bibinfo[2]{#2}
\providecommand\natexlab[1]{#1}
\providecommand\showeprint[2][]{arXiv:#2}

\bibitem[\protect\citeauthoryear{Badenes-Olmedo, Redondo-Garcia, and
  Corcho}{Badenes-Olmedo et~al\mbox{.}}{2017}]%
        {Badenes-Olmedo2017}
\bibfield{author}{\bibinfo{person}{C. Badenes-Olmedo}, \bibinfo{person}{J.L.
  Redondo-Garcia}, {and} \bibinfo{person}{O. Corcho}.}
  \bibinfo{year}{2017}\natexlab{}.
\newblock \showarticletitle{{Distributing Text Mining tasks with librAIry}}. In
  \bibinfo{booktitle}{\emph{17th ACM Symposium on Document Engineering
  (DocEng)}}.
\newblock
\showISBNx{978-1-4503-4689-4/17/09}
\urldef\tempurl%
\url{https://doi.org/10.1145/3103010.3121040}
\showDOI{\tempurl}


\bibitem[\protect\citeauthoryear{Badenes-Olmedo, Redondo-Garc{\'{i}}a, and
  Corcho}{Badenes-Olmedo et~al\mbox{.}}{2019}]%
        {Badenes-Olmedo2019}
\bibfield{author}{\bibinfo{person}{C. Badenes-Olmedo}, \bibinfo{person}{J.L.
  Redondo-Garc{\'{i}}a}, {and} \bibinfo{person}{O. Corcho}.}
  \bibinfo{year}{2019}\natexlab{}.
\newblock \showarticletitle{{Large-Scale Semantic Exploration of Scientific
  Literature using Topic-based Hashing Algorithms}}.
\newblock \bibinfo{journal}{\emph{Semantic Web Journal (under review)}}
  (\bibinfo{year}{2019}).
\newblock


\bibitem[\protect\citeauthoryear{Bagga and Baldwin}{Bagga and Baldwin}{1998}]%
        {Bagga1998}
\bibfield{author}{\bibinfo{person}{A. Bagga} {and} \bibinfo{person}{B.
  Baldwin}.} \bibinfo{year}{1998}\natexlab{}.
\newblock \showarticletitle{{Algorithms for scoring coreference chains}}. In
  \bibinfo{booktitle}{\emph{Proceedings of the 1st international conference on
  language resources and evaluation workshop on linguistics coreference}}.
  \bibinfo{pages}{563–566}.
\newblock


\bibitem[\protect\citeauthoryear{Blei, Ng, and Jordan}{Blei
  et~al\mbox{.}}{2003}]%
        {Blei2003}
\bibfield{author}{\bibinfo{person}{David~M Blei}, \bibinfo{person}{Andrew~Y
  Ng}, {and} \bibinfo{person}{Michael~I Jordan}.}
  \bibinfo{year}{2003}\natexlab{}.
\newblock \showarticletitle{{Latent Dirichlet Allocation}}.
\newblock \bibinfo{journal}{\emph{Journal of Machine Learning Research}}
  \bibinfo{volume}{3}, \bibinfo{number}{4-5} (\bibinfo{year}{2003}),
  \bibinfo{pages}{993--1022}.
\newblock
\showISBNx{9781577352815}
\showISSN{15324435}
\urldef\tempurl%
\url{https://doi.org/10.1162/jmlr.2003.3.4-5.993}
\showDOI{\tempurl}


\bibitem[\protect\citeauthoryear{Bond and Foster}{Bond and Foster}{2013}]%
        {Bond2013}
\bibfield{author}{\bibinfo{person}{Francis Bond} {and} \bibinfo{person}{Ryan
  Foster}.} \bibinfo{year}{2013}\natexlab{}.
\newblock \showarticletitle{{Linking and Extending an Open Multilingual
  Wordnet}}.
\newblock \bibinfo{journal}{\emph{Proceedings of the 51st Annual Meeting of the
  Association for Computational Linguistics (ACL)}} (\bibinfo{year}{2013}),
  \bibinfo{pages}{1352--1362}.
\newblock


\bibitem[\protect\citeauthoryear{Bond and Paik}{Bond and Paik}{2012}]%
        {Bond2012}
\bibfield{author}{\bibinfo{person}{Francis Bond} {and}
  \bibinfo{person}{Kyonghee Paik}.} \bibinfo{year}{2012}\natexlab{}.
\newblock \showarticletitle{{A survey of wordnets and their licenses}}. In
  \bibinfo{booktitle}{\emph{Proceedings of the 6th Global WordNet Conference
  (GWC 2012)}}. \bibinfo{pages}{64--71}.
\newblock


\bibitem[\protect\citeauthoryear{Cheng, Yan, Lan, and Guo}{Cheng
  et~al\mbox{.}}{2014}]%
        {Cheng2014a}
\bibfield{author}{\bibinfo{person}{Xueqi Cheng}, \bibinfo{person}{Xiaohui Yan},
  \bibinfo{person}{Yanyan Lan}, {and} \bibinfo{person}{Jiafeng Guo}.}
  \bibinfo{year}{2014}\natexlab{}.
\newblock \showarticletitle{{BTM : Topic Modeling over Short Texts}}.
\newblock \bibinfo{journal}{\emph{IEEE Transactions on Knowledge and Data
  Engineering}} \bibinfo{volume}{26}, \bibinfo{number}{12}
  (\bibinfo{year}{2014}), \bibinfo{pages}{2928--2941}.
\newblock
\urldef\tempurl%
\url{https://doi.org/10.1109/TKDE.2014.2313872}
\showDOI{\tempurl}


\bibitem[\protect\citeauthoryear{De~Smet and Moens}{De~Smet and Moens}{2009}]%
        {DeSmet2009}
\bibfield{author}{\bibinfo{person}{Wim De~Smet} {and}
  \bibinfo{person}{Marie-Francine Moens}.} \bibinfo{year}{2009}\natexlab{}.
\newblock \showarticletitle{{Cross-language linking of news stories on the web
  using interlingual topic modelling}}. In
  \bibinfo{booktitle}{\emph{Proceedings of the 2nd ACM workshop on Social web
  search and mining}}. \bibinfo{pages}{57}.
\newblock
\showISBNx{9781605588063}
\urldef\tempurl%
\url{https://doi.org/10.1145/1651437.1651447}
\showDOI{\tempurl}


\bibitem[\protect\citeauthoryear{De~Smet, Tang, and Moens}{De~Smet
  et~al\mbox{.}}{2011}]%
        {10.1007/978-3-642-20841-6_45}
\bibfield{author}{\bibinfo{person}{Wim De~Smet}, \bibinfo{person}{Jie Tang},
  {and} \bibinfo{person}{Marie-Francine Moens}.}
  \bibinfo{year}{2011}\natexlab{}.
\newblock \showarticletitle{{Knowledge Transfer across Multilingual Corpora via
  Latent Topics}}. In \bibinfo{booktitle}{\emph{Advances in Knowledge Discovery
  and Data Mining}}. \bibinfo{pages}{549--560}.
\newblock
\showISBNx{978-3-642-20841-6}


\bibitem[\protect\citeauthoryear{{Eurovoc}}{{Eurovoc}}{1995}]%
        {Eurovoc1995}
\bibfield{author}{\bibinfo{person}{{Eurovoc}}.}
  \bibinfo{year}{1995}\natexlab{}.
\newblock \showarticletitle{{Thesaurus EUROVOC - Volume 2: Subject-Oriented
  Version. Ed. 3/English Language. Annex to the index of the Official Journal
  of the EC}}. In \bibinfo{booktitle}{\emph{Luxembourg, Office for Official
  Publications of the European Communities}}.
\newblock


\bibitem[\protect\citeauthoryear{Ganguly, Leveling, and Jones}{Ganguly
  et~al\mbox{.}}{2012}]%
        {ganguly-etal-2012-cross}
\bibfield{author}{\bibinfo{person}{Debasis Ganguly}, \bibinfo{person}{Johannes
  Leveling}, {and} \bibinfo{person}{Gareth Jones}.}
  \bibinfo{year}{2012}\natexlab{}.
\newblock \showarticletitle{{Cross-Lingual Topical Relevance Models}}. In
  \bibinfo{booktitle}{\emph{Proceedings of COLING 2012}}.
  \bibinfo{pages}{927--942}.
\newblock


\bibitem[\protect\citeauthoryear{Gatti, Brooks, and Nurre}{Gatti
  et~al\mbox{.}}{2015}]%
        {Gatti2015}
\bibfield{author}{\bibinfo{person}{Christopher~J. Gatti},
  \bibinfo{person}{James~D. Brooks}, {and} \bibinfo{person}{Sarah~G. Nurre}.}
  \bibinfo{year}{2015}\natexlab{}.
\newblock \showarticletitle{{A Historical Analysis of the Field of OR/MS using
  Topic Models}}.
\newblock \bibinfo{journal}{\emph{CoRR}}  \bibinfo{volume}{abs/1510.0}
  (\bibinfo{year}{2015}).
\newblock


\bibitem[\protect\citeauthoryear{Gliozzo}{Gliozzo}{2007}]%
        {Gliozzo2007}
\bibfield{author}{\bibinfo{person}{Alfio~Massimiliano Gliozzo}.}
  \bibinfo{year}{2007}\natexlab{}.
\newblock \showarticletitle{{The Domain Restriction Hypothesis: Relating Term
  Similarity and Semantic Consistency}}.
\newblock \bibinfo{journal}{\emph{In. Proceedings of NAACL HLT}}
  \bibinfo{number}{April} (\bibinfo{year}{2007}), \bibinfo{pages}{131}.
\newblock


\bibitem[\protect\citeauthoryear{Greene and Cross}{Greene and Cross}{2016}]%
        {Greene2016}
\bibfield{author}{\bibinfo{person}{Derek Greene} {and} \bibinfo{person}{James~P
  Cross}.} \bibinfo{year}{2016}\natexlab{}.
\newblock \showarticletitle{{Exploring the political agenda of the european
  parliament using a dynamic topic modeling approach}}.
\newblock \bibinfo{journal}{\emph{Political Analysis}} \bibinfo{volume}{25},
  \bibinfo{number}{1} (\bibinfo{year}{2016}), \bibinfo{pages}{77--94}.
\newblock
\showISBNx{9781450336727}
\showISSN{14764989}
\urldef\tempurl%
\url{https://doi.org/10.1017/pan.2016.7}
\showDOI{\tempurl}


\bibitem[\protect\citeauthoryear{Griffiths and Steyvers}{Griffiths and
  Steyvers}{2004}]%
        {Griffiths2004b}
\bibfield{author}{\bibinfo{person}{Thomas~L Griffiths} {and}
  \bibinfo{person}{Mark Steyvers}.} \bibinfo{year}{2004}\natexlab{}.
\newblock \showarticletitle{{Finding scientific topics.}}
\newblock \bibinfo{journal}{\emph{Proceedings of the National Academy of
  Sciences of the United States of America}}  \bibinfo{volume}{101 Suppl}
  (\bibinfo{year}{2004}), \bibinfo{pages}{5228--35}.
\newblock
\showISSN{0027-8424}
\urldef\tempurl%
\url{https://doi.org/10.1073/pnas.0307752101}
\showDOI{\tempurl}


\bibitem[\protect\citeauthoryear{Gutierrez, Shutova, Lichtenstein, Melo, and
  Gilardi}{Gutierrez et~al\mbox{.}}{2016}]%
        {errez2016}
\bibfield{author}{\bibinfo{person}{E.~Dario Gutierrez},
  \bibinfo{person}{Ekaterina Shutova}, \bibinfo{person}{Patricia Lichtenstein},
  \bibinfo{person}{Gerard~de Melo}, {and} \bibinfo{person}{Luca Gilardi}.}
  \bibinfo{year}{2016}\natexlab{}.
\newblock \showarticletitle{{Detecting Cross-cultural Differences Using a
  Multilingual Topic Model}}.
\newblock \bibinfo{journal}{\emph{Transactions of the Association for
  Computational Linguistics}}  \bibinfo{volume}{4} (\bibinfo{year}{2016}),
  \bibinfo{pages}{47--60}.
\newblock


\bibitem[\protect\citeauthoryear{Hao, Boyd-Graber, and Paul.}{Hao
  et~al\mbox{.}}{2018}]%
        {Hao2018}
\bibfield{author}{\bibinfo{person}{Shudong Hao}, \bibinfo{person}{Jordan~L.
  Boyd-Graber}, {and} \bibinfo{person}{Michael~J. Paul.}}
  \bibinfo{year}{2018}\natexlab{}.
\newblock \showarticletitle{{Lessons from the Bible on Modern Topics: Adapting
  Topic Model Evaluation to Multilingual and Low-Resource Settings}}. In
  \bibinfo{booktitle}{\emph{Proceedings of the 2018 Conference of the North
  American Chapter of the Association for Computational Linguis- tics: Human
  Language Technologies, NAACL-HLT}}. \bibinfo{pages}{1090–1100}.
\newblock


\bibitem[\protect\citeauthoryear{Hao and Paul}{Hao and Paul}{2018}]%
        {Hao2018b}
\bibfield{author}{\bibinfo{person}{Shudong Hao} {and}
  \bibinfo{person}{Michael~J. Paul}.} \bibinfo{year}{2018}\natexlab{}.
\newblock \showarticletitle{{Learning Multilingual Topics from Incomparable
  Corpus}}. In \bibinfo{booktitle}{\emph{COLING}}.
\newblock


\bibitem[\protect\citeauthoryear{He, Li, and Wu}{He et~al\mbox{.}}{2017}]%
        {He2017}
\bibfield{author}{\bibinfo{person}{Jin He}, \bibinfo{person}{Lei Li}, {and}
  \bibinfo{person}{Xindong Wu}.} \bibinfo{year}{2017}\natexlab{}.
\newblock \showarticletitle{{A self-adaptive sliding window based topic model
  for non-uniform texts}}. In \bibinfo{booktitle}{\emph{Proceedings - IEEE
  International Conference on Data Mining, ICDM}},
  Vol.~\bibinfo{volume}{2017-Novem}. \bibinfo{pages}{147--156}.
\newblock
\showISBNx{9781538638347}
\showISSN{15504786}
\urldef\tempurl%
\url{https://doi.org/10.1109/ICDM.2017.24}
\showDOI{\tempurl}


\bibitem[\protect\citeauthoryear{Hu, Zhai, Eidelman, and Boyd-Graber.}{Hu
  et~al\mbox{.}}{2014}]%
        {Hu2014a}
\bibfield{author}{\bibinfo{person}{Yuening Hu}, \bibinfo{person}{Ke Zhai},
  \bibinfo{person}{Vladimir Eidelman}, {and} \bibinfo{person}{Jordan~L.
  Boyd-Graber.}} \bibinfo{year}{2014}\natexlab{}.
\newblock \showarticletitle{{Polylingual Tree-Based Topic Models for
  Translation Domain Adaptation}}. In \bibinfo{booktitle}{\emph{Proceedings of
  the 52nd Annual Meeting of the Association for Computational Linguistics}}.
  \bibinfo{pages}{1166–1176}.
\newblock


\bibitem[\protect\citeauthoryear{Ji, Li, Yan, Tian, and Zhang}{Ji
  et~al\mbox{.}}{2013}]%
        {Ji2013}
\bibfield{author}{\bibinfo{person}{Jianqiu Ji}, \bibinfo{person}{Jianmin Li},
  \bibinfo{person}{Shuicheng Yan}, \bibinfo{person}{Qi Tian}, {and}
  \bibinfo{person}{Bo Zhang}.} \bibinfo{year}{2013}\natexlab{}.
\newblock \showarticletitle{{Min-Max Hash for Jaccard Similarity}}. In
  \bibinfo{booktitle}{\emph{2013 IEEE 13th International Conference on Data
  Mining}}. \bibinfo{publisher}{IEEE}, \bibinfo{pages}{301--309}.
\newblock
\showISBNx{978-0-7695-5108-1}
\urldef\tempurl%
\url{https://doi.org/10.1109/ICDM.2013.119}
\showDOI{\tempurl}


\bibitem[\protect\citeauthoryear{Li and K{\"{o}}nig}{Li and
  K{\"{o}}nig}{2010}]%
        {Li2010b}
\bibfield{author}{\bibinfo{person}{Ping Li} {and} \bibinfo{person}{Christian
  K{\"{o}}nig}.} \bibinfo{year}{2010}\natexlab{}.
\newblock \showarticletitle{{b-Bit minwise hashing}}. In
  \bibinfo{booktitle}{\emph{Proceedings of the 19th international conference on
  World wide web - WWW '10}}. \bibinfo{publisher}{ACM Press},
  \bibinfo{pages}{671}.
\newblock
\showISBNx{9781605587998}
\urldef\tempurl%
\url{https://doi.org/10.1145/1772690.1772759}
\showDOI{\tempurl}


\bibitem[\protect\citeauthoryear{Li, Owen, and Zhang}{Li et~al\mbox{.}}{2012}]%
        {Li2012}
\bibfield{author}{\bibinfo{person}{Ping Li}, \bibinfo{person}{Art~B. Owen},
  {and} \bibinfo{person}{Cun-Hui Zhang}.} \bibinfo{year}{2012}\natexlab{}.
\newblock \showarticletitle{{One Permutation Hashing}}.
\newblock \bibinfo{journal}{\emph{Advances in Neural Information Processing}}
  (\bibinfo{year}{2012}).
\newblock


\bibitem[\protect\citeauthoryear{Lu, Wei, and Hsiao}{Lu et~al\mbox{.}}{2016}]%
        {Lu2016}
\bibfield{author}{\bibinfo{person}{Hsin~Min Lu}, \bibinfo{person}{Chih~Ping
  Wei}, {and} \bibinfo{person}{Fei~Yuan Hsiao}.}
  \bibinfo{year}{2016}\natexlab{}.
\newblock \showarticletitle{{Modeling healthcare data using multiple-channel
  latent Dirichlet allocation}}.
\newblock \bibinfo{journal}{\emph{Journal of Biomedical Informatics}}
  \bibinfo{volume}{60} (\bibinfo{year}{2016}), \bibinfo{pages}{210--223}.
\newblock
\showISSN{15320464}
\urldef\tempurl%
\url{https://doi.org/10.1016/j.jbi.2016.02.003}
\showDOI{\tempurl}


\bibitem[\protect\citeauthoryear{Ma and Nasukawa}{Ma and Nasukawa}{2017}]%
        {Ma2017}
\bibfield{author}{\bibinfo{person}{Tengfei Ma} {and} \bibinfo{person}{Tetsuya
  Nasukawa}.} \bibinfo{year}{2017}\natexlab{}.
\newblock \showarticletitle{{Inverted Bilingual Topic Models for Lexicon
  Extraction from Non-parallel Data}}. In \bibinfo{booktitle}{\emph{Proceedings
  of the Twenty-Sixth International Joint Conference on Artificial
  Intelligence}}. \bibinfo{pages}{4075--4081}.
\newblock


\bibitem[\protect\citeauthoryear{Mao, Feng, Hao, Nie, Huang, and Wen}{Mao
  et~al\mbox{.}}{2017}]%
        {Mao2017}
\bibfield{author}{\bibinfo{person}{Xianling Mao}, \bibinfo{person}{Bo-Si Feng},
  \bibinfo{person}{Yi-Jing Hao}, \bibinfo{person}{Liqiang Nie},
  \bibinfo{person}{Heyan Huang}, {and} \bibinfo{person}{Guihua Wen}.}
  \bibinfo{year}{2017}\natexlab{}.
\newblock \showarticletitle{{S2JSD-LSH: A Locality-Sensitive Hashing Schema for
  Probability Distributions}}. In \bibinfo{booktitle}{\emph{AAAI}}.
\newblock


\bibitem[\protect\citeauthoryear{Miller}{Miller}{1995}]%
        {Miller1995WordNet:English}
\bibfield{author}{\bibinfo{person}{George~A. Miller}.}
  \bibinfo{year}{1995}\natexlab{}.
\newblock \showarticletitle{{WordNet: A Lexical Database for English}}.
\newblock \bibinfo{journal}{\emph{Commun. ACM}} \bibinfo{volume}{38},
  \bibinfo{number}{11} (\bibinfo{year}{1995}), \bibinfo{pages}{39--41}.
\newblock


\bibitem[\protect\citeauthoryear{Mimno, Wallach, Talley, Leenders, and
  McCallum}{Mimno et~al\mbox{.}}{2011}]%
        {mimno-etal-2011-optimizing}
\bibfield{author}{\bibinfo{person}{David Mimno}, \bibinfo{person}{Hanna
  Wallach}, \bibinfo{person}{Edmund Talley}, \bibinfo{person}{Miriam Leenders},
  {and} \bibinfo{person}{Andrew McCallum}.} \bibinfo{year}{2011}\natexlab{}.
\newblock \showarticletitle{{Optimizing Semantic Coherence in Topic Models}}.
  In \bibinfo{booktitle}{\emph{Proceedings of the 2011 Conference on Empirical
  Methods in Natural Language Processing}}. \bibinfo{publisher}{Association for
  Computational Linguistics}, \bibinfo{address}{Edinburgh, Scotland, UK.},
  \bibinfo{pages}{262--272}.
\newblock


\bibitem[\protect\citeauthoryear{Mimno, Wallach, Naradowsky, Smith, and
  McCallum}{Mimno et~al\mbox{.}}{2009}]%
        {Mimno:2009:PTM:1699571.1699627}
\bibfield{author}{\bibinfo{person}{David Mimno}, \bibinfo{person}{Hanna~M
  Wallach}, \bibinfo{person}{Jason Naradowsky}, \bibinfo{person}{David~A
  Smith}, {and} \bibinfo{person}{Andrew McCallum}.}
  \bibinfo{year}{2009}\natexlab{}.
\newblock \showarticletitle{{Polylingual Topic Models}}. In
  \bibinfo{booktitle}{\emph{Proceedings of the 2009 Conference on Empirical
  Methods in Natural Language Processing: Volume 2 - Volume 2}}
  \emph{(\bibinfo{series}{EMNLP '09})}. \bibinfo{publisher}{Association for
  Computational Linguistics}, \bibinfo{address}{Stroudsburg, PA, USA},
  \bibinfo{pages}{880--889}.
\newblock
\showISBNx{978-1-932432-62-6}


\bibitem[\protect\citeauthoryear{Moritz and Bunchler}{Moritz and
  Bunchler}{2017}]%
        {Moritz2017}
\bibfield{author}{\bibinfo{person}{Maria Moritz} {and} \bibinfo{person}{Marco
  Bunchler}.} \bibinfo{year}{2017}\natexlab{}.
\newblock \showarticletitle{{Ambiguity in Semantically Related Word
  Substitutions: an Investigation in Historical Bible Translations.}}. In
  \bibinfo{booktitle}{\emph{Proceedings of the NoDaLiDa 2017 Workshop on
  Processing Historical Language}}. \bibinfo{pages}{18--23}.
\newblock


\bibitem[\protect\citeauthoryear{Newman, Lau, Grieser, and Baldwin}{Newman
  et~al\mbox{.}}{2010}]%
        {newman-etal-2010-automatic}
\bibfield{author}{\bibinfo{person}{David Newman}, \bibinfo{person}{Jey~Han
  Lau}, \bibinfo{person}{Karl Grieser}, {and} \bibinfo{person}{Timothy
  Baldwin}.} \bibinfo{year}{2010}\natexlab{}.
\newblock \showarticletitle{{Automatic Evaluation of Topic Coherence}}. In
  \bibinfo{booktitle}{\emph{Human Language Technologies: The 2010 Annual
  Conference of the North {\{}A{\}}merican Chapter of the Association for
  Computational Linguistics}}. \bibinfo{pages}{100--108}.
\newblock


\bibitem[\protect\citeauthoryear{Ni, Sun, Hu, and Chen}{Ni
  et~al\mbox{.}}{2011}]%
        {Ni:2011:CLT:1935826.1935887}
\bibfield{author}{\bibinfo{person}{Xiaochuan Ni}, \bibinfo{person}{Jian-Tao
  Sun}, \bibinfo{person}{Jian Hu}, {and} \bibinfo{person}{Zheng Chen}.}
  \bibinfo{year}{2011}\natexlab{}.
\newblock \showarticletitle{{Cross Lingual Text Classification by Mining
  Multilingual Topics from Wikipedia}}. In
  \bibinfo{booktitle}{\emph{Proceedings of the Fourth ACM International
  Conference on Web Search and Data Mining}}. \bibinfo{pages}{375--384}.
\newblock
\showISBNx{978-1-4503-0493-1}
\urldef\tempurl%
\url{https://doi.org/10.1145/1935826.1935887}
\showDOI{\tempurl}


\bibitem[\protect\citeauthoryear{O’Neill, Robin, O’Brien, and
  Buitelaar}{O’Neill et~al\mbox{.}}{2017}]%
        {ONeill2017}
\bibfield{author}{\bibinfo{person}{James O’Neill}, \bibinfo{person}{Cécile
  Robin}, \bibinfo{person}{Leona O’Brien}, {and} \bibinfo{person}{Paul
  Buitelaar}.} \bibinfo{year}{2017}\natexlab{}.
\newblock \showarticletitle{{An analysis of topic modelling for legislative
  texts}}.
\newblock \bibinfo{journal}{\emph{CEUR Workshop Proceedings}}
  \bibinfo{volume}{2143} (\bibinfo{year}{2017}).
\newblock
\showISBNx{1234567245}
\showISSN{16130073}
\urldef\tempurl%
\url{https://doi.org/10.475/123}
\showDOI{\tempurl}


\bibitem[\protect\citeauthoryear{Platt, Toutanova, and Yih}{Platt
  et~al\mbox{.}}{2010}]%
        {Platt:2010:TDR:1870658.1870683}
\bibfield{author}{\bibinfo{person}{John~C Platt}, \bibinfo{person}{Kristina
  Toutanova}, {and} \bibinfo{person}{Wen-tau Yih}.}
  \bibinfo{year}{2010}\natexlab{}.
\newblock \showarticletitle{{Translingual Document Representations from
  Discriminative Projections}}. In \bibinfo{booktitle}{\emph{Proceedings of the
  2010 Conference on Empirical Methods in Natural Language Processing}}
  \emph{(\bibinfo{series}{EMNLP '10})}. \bibinfo{publisher}{Association for
  Computational Linguistics}, \bibinfo{address}{Stroudsburg, PA, USA},
  \bibinfo{pages}{251--261}.
\newblock


\bibitem[\protect\citeauthoryear{Ramage, Dumais, and Liebling}{Ramage
  et~al\mbox{.}}{2010}]%
        {Ramage2010}
\bibfield{author}{\bibinfo{person}{Daniel Ramage}, \bibinfo{person}{Susan
  Dumais}, {and} \bibinfo{person}{Dan Liebling}.}
  \bibinfo{year}{2010}\natexlab{}.
\newblock \showarticletitle{{Characterizing Microblogs with Topic Models}}. In
  \bibinfo{booktitle}{\emph{Proceedings of the Fourth International Conference
  on Weblogs and Social Media}}. \bibinfo{pages}{1--8}.
\newblock
\showISBNx{9781577354451}


\bibitem[\protect\citeauthoryear{Ramage, Hall, Nallapati, and Manning}{Ramage
  et~al\mbox{.}}{2009}]%
        {Ramage2009a}
\bibfield{author}{\bibinfo{person}{Daniel Ramage}, \bibinfo{person}{David
  Hall}, \bibinfo{person}{Ramesh Nallapati}, {and}
  \bibinfo{person}{Christopher~D Manning}.} \bibinfo{year}{2009}\natexlab{}.
\newblock \showarticletitle{{Labeled LDA: A supervised topic model for credit
  attribution in multi-labeled corpora}}. In
  \bibinfo{booktitle}{\emph{Proceedings of the 2009 Conference on Empirical
  Methods in Natural Language Processing}}. \bibinfo{pages}{248--256}.
\newblock
\showISBNx{978-1-932432-59-6}
\showISSN{1932432590}
\urldef\tempurl%
\url{https://doi.org/10.3115/1699510.1699543}
\showDOI{\tempurl}


\bibitem[\protect\citeauthoryear{Steinberger, Pouliquen, Widiger, Ignat,
  Erjavec, Tufis, and Varga}{Steinberger et~al\mbox{.}}{2006}]%
        {Steinberger2006}
\bibfield{author}{\bibinfo{person}{Ralf Steinberger}, \bibinfo{person}{Bruno
  Pouliquen}, \bibinfo{person}{Anna Widiger}, \bibinfo{person}{Camelia Ignat},
  \bibinfo{person}{Tomaz Erjavec}, \bibinfo{person}{Dan Tufis}, {and}
  \bibinfo{person}{Daniel Varga}.} \bibinfo{year}{2006}\natexlab{}.
\newblock \showarticletitle{{The JRC-Acquis: A multilingual aligned parallel
  corpus with 20+ languages}}. In \bibinfo{booktitle}{\emph{Proceedings of the
  5th International Conference on Language Resources and Evaluation
  (LREC'2006)}}, Vol.~\bibinfo{volume}{4}. \bibinfo{pages}{2142--2147}.
\newblock


\bibitem[\protect\citeauthoryear{Tapi~Nzali, Bringay, Lavergne, Mollevi, and
  Opitz}{Tapi~Nzali et~al\mbox{.}}{2017}]%
        {TapiNzali2017}
\bibfield{author}{\bibinfo{person}{Mike~Donald Tapi~Nzali},
  \bibinfo{person}{Sandra Bringay}, \bibinfo{person}{Christian Lavergne},
  \bibinfo{person}{Caroline Mollevi}, {and} \bibinfo{person}{Thomas Opitz}.}
  \bibinfo{year}{2017}\natexlab{}.
\newblock \showarticletitle{{What Patients Can Tell Us: Topic Analysis for
  Social Media on Breast Cancer.}}
\newblock \bibinfo{journal}{\emph{JMIR medical informatics}}
  \bibinfo{volume}{5}, \bibinfo{number}{3} (\bibinfo{date}{7}
  \bibinfo{year}{2017}), \bibinfo{pages}{e23}.
\newblock
\showISSN{2291-9694}
\urldef\tempurl%
\url{https://doi.org/10.2196/medinform.7779}
\showDOI{\tempurl}


\bibitem[\protect\citeauthoryear{Vuli{\'{c}}, De~Smet, Tang, and
  Moens}{Vuli{\'{c}} et~al\mbox{.}}{2015}]%
        {Vulic2015}
\bibfield{author}{\bibinfo{person}{Ivan Vuli{\'{c}}}, \bibinfo{person}{Wim
  De~Smet}, \bibinfo{person}{Jie Tang}, {and} \bibinfo{person}{Marie~Francine
  Moens}.} \bibinfo{year}{2015}\natexlab{}.
\newblock \showarticletitle{{Probabilistic topic modeling in multilingual
  settings: An overview of its methodology and applications}}.
\newblock \bibinfo{journal}{\emph{Information Processing and Management}}
  \bibinfo{volume}{51}, \bibinfo{number}{1} (\bibinfo{year}{2015}),
  \bibinfo{pages}{111--147}.
\newblock
\showISBNx{0306-4573}
\showISSN{03064573}
\urldef\tempurl%
\url{https://doi.org/10.1016/j.ipm.2014.08.003}
\showDOI{\tempurl}


\bibitem[\protect\citeauthoryear{Vuli{\'{c}} and Moens}{Vuli{\'{c}} and
  Moens}{2012}]%
        {Vulic:2012:DHC:2380816.2380872}
\bibfield{author}{\bibinfo{person}{Ivan Vuli{\'{c}}} {and}
  \bibinfo{person}{Marie-Francine Moens}.} \bibinfo{year}{2012}\natexlab{}.
\newblock \showarticletitle{{Detecting Highly Confident Word Translations from
  Comparable Corpora Without Any Prior Knowledge}}. In
  \bibinfo{booktitle}{\emph{Proceedings of the 13th Conference of the European
  Chapter of the Association for Computational Linguistics}}.
  \bibinfo{pages}{449--459}.
\newblock
\showISBNx{978-1-937284-19-0}


\bibitem[\protect\citeauthoryear{Vuli{\'{c}} and Moens}{Vuli{\'{c}} and
  Moens}{2013}]%
        {10.1007/978-3-642-36973-5_9}
\bibfield{author}{\bibinfo{person}{Ivan Vuli{\'{c}}} {and}
  \bibinfo{person}{Marie-Francine Moens}.} \bibinfo{year}{2013}\natexlab{}.
\newblock \showarticletitle{{A Unified Framework for Monolingual and
  Cross-Lingual Relevance Modeling Based on Probabilistic Topic Models}}. In
  \bibinfo{booktitle}{\emph{Advances in Information Retrieval}}.
  \bibinfo{pages}{98--109}.
\newblock
\showISBNx{978-3-642-36973-5}


\bibitem[\protect\citeauthoryear{Zhao, J{\'{e}}gou, and Gravier}{Zhao
  et~al\mbox{.}}{2013}]%
        {Zhao2013}
\bibfield{author}{\bibinfo{person}{Wan-Lei Zhao}, \bibinfo{person}{Hervé
  J{\'{e}}gou}, {and} \bibinfo{person}{Guillaume Gravier}.}
  \bibinfo{year}{2013}\natexlab{}.
\newblock \showarticletitle{{Sim-min-hash}}. In
  \bibinfo{booktitle}{\emph{Proceedings of the 21st ACM international
  conference on Multimedia - MM '13}}. \bibinfo{pages}{577--580}.
\newblock
\showISBNx{9781450324045}
\urldef\tempurl%
\url{https://doi.org/10.1145/2502081.2502152}
\showDOI{\tempurl}


\bibitem[\protect\citeauthoryear{Zhen, Gao, Yeung, Zha, and Li}{Zhen
  et~al\mbox{.}}{2016}]%
        {Zhen2016}
\bibfield{author}{\bibinfo{person}{Yi Zhen}, \bibinfo{person}{Yue Gao},
  \bibinfo{person}{Dit-Yan Yeung}, \bibinfo{person}{Hongyuan Zha}, {and}
  \bibinfo{person}{Xuelong Li}.} \bibinfo{year}{2016}\natexlab{}.
\newblock \showarticletitle{{Spectral Multimodal Hashing and Its Application to
  Multimedia Retrieval}}.
\newblock \bibinfo{journal}{\emph{IEEE Transactions on Cybernetics}}
  \bibinfo{volume}{46}, \bibinfo{number}{1} (\bibinfo{year}{2016}),
  \bibinfo{pages}{27--38}.
\newblock
\showISSN{2168-2267}
\urldef\tempurl%
\url{https://doi.org/10.1109/TCYB.2015.2392052}
\showDOI{\tempurl}


\bibitem[\protect\citeauthoryear{Zhu, Li, Chen, and Yang}{Zhu
  et~al\mbox{.}}{2013}]%
        {zhu-etal-2013-building}
\bibfield{author}{\bibinfo{person}{Zede Zhu}, \bibinfo{person}{Miao Li},
  \bibinfo{person}{Lei Chen}, {and} \bibinfo{person}{Zhenxin Yang}.}
  \bibinfo{year}{2013}\natexlab{}.
\newblock \showarticletitle{{Building Comparable Corpora Based on Bilingual
  {\{}LDA{\}} Model}}. In \bibinfo{booktitle}{\emph{Proceedings of the 51st
  Annual Meeting of the Association for Computational Linguistics (Volume 2:
  Short Papers)}}. \bibinfo{pages}{278--282}.
\newblock


\end{thebibliography}

\end{document}